\documentclass[journal]{IEEEtran}

\usepackage{cite}
\usepackage{amsmath,amssymb,amsfonts}
\usepackage{algorithm}
\usepackage{algorithmic}
\usepackage{graphicx}
\usepackage{textcomp}
\usepackage{multirow}
\usepackage{threeparttable}
\usepackage[table,xcdraw]{xcolor}
\usepackage{subfigure}
\usepackage{epstopdf}
\usepackage{changepage}
\usepackage{courier}
\usepackage{cooltooltips}

\newcommand{\xt}{\mathbf{x}}
\newcommand{\ft}{\mathbf{f}}

\newcommand{\Pt}{\mathbf{P}}
\newcommand{\Qt}{\mathbf{Q}}
\newcommand{\At}{\mathbf{A}}
\newcommand{\at}{\mathbf{a}}

\newcommand{\zt}{\mathbf{z}}

\begin{document}

\title{Evolutionary Robust Clustering Over Time for Temporal Data}

\author{Qi~Zhao,
	Bai~Yan, and
	Yuhui~Shi,~\IEEEmembership{Fellow,~IEEE}
	
	\thanks{This paper is supported by the National Natural Science Foundation of China under Grant 61761136008, Science and Technology Innovation Committee Foundation of Shenzhen under Grant JCYJ20200109141235597, Shenzhen Peacock Plan under Grant KQTD2016112514355531, and Program for Guangdong Introducing Innovative and Entrepreneurial Teams under Grant 2017ZT07X386. \textit{(Corresponding author: Yuhui Shi)}}
	
	\thanks{Q. Zhao and B. Yan are with the Department of Computer Science and Engineering, Southern University of Science and Technology, Shenzhen 518055, China, and also with the School of Computer Science and Technology, University of Science and Technology of China, Hefei 230027, China (e-mails: zhaoq@sustech.edu.cn; yanb@sustech.edu.cn).}
	
	\thanks{Y. Shi is with the Department of Computer Science and Engineering, Southern University of Science and Technology, Shenzhen 518055, China (e-mail: shiyh@sustech.edu.cn).}
	
	}

\maketitle

\begin{abstract}
In many clustering scenes, data samples' attribute values change over time. For such data, we are often interested in obtaining a partition for each time step and tracking the dynamic change of partitions. Normally, a smooth change is assumed for data to have a temporal smooth nature. Existing algorithms consider the temporal smoothness as an \emph{a priori} preference and bias the search towards the preferred direction. This \emph{a priori} manner leads to a risk of converging to an unexpected region because it is not always the case that a reasonable preference can be elicited given the little prior knowledge about the data. To address this issue, this paper proposes a new clustering framework called evolutionary robust clustering over time. One significant innovation of the proposed framework is processing the temporal smoothness in an \emph{a posteriori} manner, which avoids unexpected convergence that occurs in existing algorithms. Furthermore, the proposed framework automatically tunes the weight of smoothness without data's affinity matrix and predefined parameters, which holds better applicability and scalability. The effectiveness and efficiency of the proposed framework are confirmed by comparing with state-of-the-art algorithms on both synthetic and real datasets.
\end{abstract}

\begin{IEEEkeywords}
evolutionary algorithm; dynamic optimization; evolutionary clustering; temporal data.
\end{IEEEkeywords}

\IEEEpeerreviewmaketitle

\section{Introduction}\label{secIntro}
\IEEEPARstart{T}{emporal} data clustering refers to partitioning data samples whose attribute values change over time \cite{xu2014adaptive}. It requires 1) obtaining a partition for each time step, and 2) tracking the temporal change of partitions. Normally, the change is expected to be smooth over time with the assumption of temporal smooth nature of data \cite{chakrabarti2006evolutionary}. For example, in video segmentation tasks, segmentation can be conducted frame by frame, and results of successive frames have temporal continuity because adjacent frames share similarities \cite{chacon2018bio}\cite{xia2018human}. Such instances are also ubiquitous in dynamic community detection \cite{folino2014evolutionary}\cite{bara2016new}\cite{zeng2019consensus}, user preference mining \cite{rana2014evolutionary}, and consumer behavior analysis \cite{lin2019opec}, etc. 

Static clustering that processes data at each time step independently is inappropriate for temporal data, because it is susceptible to noise, and the obtained partitions are unstable at neighboring time periods. An alternative way is to incorporate a temporal smoothness penalty with static clustering approaches. This penalty guarantees the partition evolving smoothly over time. Algorithms following this way can be divided into two types. 

The first type incorporates temporal smoothness penalty into clustering validity indexes \cite{chakrabarti2006evolutionary}\cite{chen2014evolutionary}\cite{li2015sparsity}\cite{langone2016efficient}\cite{wang2017hierarchical} or considers the penalty when calculating the affinity matrix/representation of data \cite{xu2014adaptive}\cite{chi2009evolutionary}\cite{hashemi2018evolutionary}\cite{arzeno2017evolutionary}\cite{arzeno2019evolutionary}. In \emph{online} clustering scenarios \cite{chi2009evolutionary}, the weight of the penalty determines to what extend the current partition/affinity matrix/representation is similar to previous ones. Most algorithms preset the weight before clustering \cite{chakrabarti2006evolutionary}\cite{chen2014evolutionary}\cite{li2015sparsity}\cite{langone2016efficient}\cite{wang2017hierarchical}\cite{chi2009evolutionary}\cite{arzeno2017evolutionary}\cite{arzeno2019evolutionary}. This setting is inapplicable in practice since the amount of temporal changes in data is unavailable a prior. A few algorithms automatically tune the weight by quantifying the significance of previous data on current data clustering \cite{xu2014adaptive}\cite{hashemi2018evolutionary}. The former \cite{xu2014adaptive} involves data's affinity matrix, in which the quadratic computational complexity and memory demand of the affinity matrix, i.e., $\mathcal{O}(S^2)$, may hinder its scalability to datasets with a large number $S$ of samples \cite{langone2016efficient}. The latter \cite{hashemi2018evolutionary} is specific to subspace clustering.                    

The second type simultaneously optimizes the clustering validity index and temporal smoothness penalty by evolutionary\footnote{In this paper, the term "evolutionary" refers to the algorithms inspired by biological evolution (i.e., evolutionary algorithms (EAs)) rather than the temporal evolution of data.} multiobjective optimization (EMO) algorithms without tuning the weight  \cite{folino2014evolutionary}\cite{bara2016new}. The smoothness penalty in \cite{folino2014evolutionary}\cite{bara2016new} is denoted by the normalized mutual information (NMI) \cite{strehl2002cluster} between the current and historical partitions. This denotation is improper to cases with changes in data samples' memberships and may lead to error propagation, i.e., historical partition error misleads the current clustering. Furthermore, these methods \cite{folino2014evolutionary}\cite{bara2016new} are tailored to dynamic community detection, being impracticable for other tasks. 

Besides, both of the two types treat the temporal smoothness as an \emph{a priori} preference, i.e., incorporating smoothness within the clustering process and searching solutions (i.e., partitions of the dataset) along with the ``preferred" direction. This \emph{a priori} manner may make algorithms converge to an unexpected region because it is not rare to struggle to set a reasonable preference given the little prior knowledge about the data \cite{li2020does}.

To address these issues, this paper proposes a new clustering framework called evolutionary robust\footnote{In this paper, the term "robust" refers to the stability of the clustering result in the time space \cite{fu2015robust} instead of resisting uncertainties in the parameter space.} clustering over time (ERCOT). ERCOT offers two innovations to temporal data clustering: 1) ERCOT processes the temporal smoothness as an \emph{a posteriori} preference, avoiding the potential unexpected convergence in existing methods. At each time step, ERCOT first optimizes the clustering objective function without considering the temporal smoothness penalty, which promotes the global convergence concerning the current data. By utilizing EA's population-based search property, multiple solutions with different smoothness concerning historical data are obtained. Solutions with appropriate smoothness can subsequently be selected \emph{a posteriori}, maintaining the partition's temporal smoothness. 2) In the procedure of selecting solutions with appropriate smoothness, ERCOT automatically tunes the weight of smoothness without data's affinity matrix, guaranteeing ERCOT's scalability. The proposed weight auto-tuning method derives the probabilistic models of data distribution at adjacent time periods and automatically infers the weights of data via model stacking. Overall, this paper's main contributions are:
\begin{itemize} 
	\item ERCOT. The proposed framework automatically processes the temporal smoothness in an \emph{a posteriori} manner, which avoids unexpected convergence that occurs in existing algorithms. 
	\item Smoothness weight auto-tuning method. This method calculates the temporal smoothness weight for ERCOT without predefined parameters and data's affinity matrix, which holds better scalability than existing algorithms.   
	\item Empirical validation of ERCOT's clustering accuracy and time efficiency on both synthetic and real datasets. Simulation results demonstrate the superiority of ERCOT with comparison to representative temporal data clustering algorithms. 
\end{itemize}

In this paper, bold capital and bold lowercase letters represent matrices and vectors, respectively. $\at_s$ is the $s$th row of $\At$.  $|\At|$ denotes the number of rows of $\At$. $\At_i$ is a sub-matrix of $\At$ with rows (data samples) belonging to the $i$th cluster. $\zt_c$ is a vector representing the $c$th cluster centroid, which can be extracted from $\zt$. $\alpha^t,\at^t,\At^t$ refer to the evolving scalar, vector, and matrix at time $t$, respectively. $\Pt^{G,t}$ denotes the evolving matrix in the $G$th algorithmic generation of time $t$.  

The remainder of the paper contains: literature review in Section \ref{secLiterature}, the proposed ERCOT in Section \ref{secAlgorithm},  experiments in Section \ref{secExp}, and conclusions in Section \ref{secConclusion}.

\section{Related Work}\label{secLiterature}
\subsection{Temporal Data Clustering}\label{sec_clustering_review}
Temporal data clustering handles data samples whose attribute values change over time. The focus is to obtain a partition for each time step and keep the partition's smoothness over consecutive time steps. This clustering task is conceptually different from data stream clustering and incremental clustering. In data stream clustering, data samples' attribute values remain unchanged, but new samples continually appear as time goes by. The aim is to partition data in a single pass with restricted memory. Incremental clustering could handle temporal data, but it does not concern about temporal smoothness but low computational cost at the price of clustering accuracy \cite{xu2014adaptive}.  

To jointly consider clustering accuracy and temporal smoothness, the seminal work \cite{chakrabarti2006evolutionary} added a smoothness penalty to the objective function of \emph{k}-means \cite{hartigan1979algorithm} and agglomerative hierarchical clustering. The objective function \cite{chakrabarti2006evolutionary} is 
\begin{equation}
J_{total}=(1-\alpha)J_{snapshot}+\alpha J_{temporal},
\end{equation}
where the total objective $J_{total}$ is an aggregation of the present clustering accuracy $J_{snapshot}$ and temporal smoothness $J_{temporal}$, and $\alpha$ is a user-defined parameter determining the weight of smoothness. Chi \emph{et al.} \cite{chi2009evolutionary} extended the above work to spectral clustering and proposed two smoothness measures, i.e., preserving cluster quality and preserving cluster membership. The former evaluates the quality of the current partition in clustering historical data, whereas the latter estimates the similarity between the present and historical partitions. The temporal smoothness was further considered in co-clustering methods \cite{li2015sparsity}, and the computational complexity of spectral clustering was reduced by the E$^2$SC algorithm \cite{langone2016efficient}. Later, the affinity propagation was introduced to deal with temporal smoothness by exchanging messages among the data at different time points \cite{arzeno2017evolutionary}\cite{arzeno2019evolutionary}. Dirichlet process was also employed for modeling the time-varying clusters, in which the partition was inferred via Gibbs sampling \cite{wang2017hierarchical}. 

The above algorithms need to predefine the value of $\alpha$, which is nontrivial in practice because the amount of temporal changes in data is unavailable a prior. To address this issue, the AFFECT \cite{xu2014adaptive} adaptively estimated the optimal weight of smoothness by shrinkage estimation. The CESM \cite{hashemi2018evolutionary} proposed a data self-expressive model for temporal data clustering in subspaces. The weight of smoothness was automatically inferred by alternately minimizing the model. 

In \cite{ma2011spatio}, the clustering validity index and temporal smoothness were simultaneously optimized by EMO without tuning the weight. This idea was later employed in dynamic community detection \cite{folino2014evolutionary}\cite{bara2016new}. These methods \cite{folino2014evolutionary}\cite{bara2016new}\cite{ma2011spatio}, together with the approach proposed in \cite{chen2015clustering}, quantify the temporal smoothness either by the distance between present and historical cluster centroids or by the NMI \cite{strehl2002cluster} between present and historical partitions. This quantification may result in error propagation, i.e., the error in historical cluster centers/partition may mislead the current clustering. Additionally, almost all the existing methods \cite{gao2016clustering} consider the temporal smoothness as an \emph{a priori} preference, i.e., inserting smoothness into the clustering process and restricting the search of solutions along with the ``preferred" direction. This \emph{a priori} manner leads to a risk of losing solution diversity and converging to an unexpected region since it is not always the case that a reasonable preference can be elicited given the little prior knowledge about the data \cite{li2020does}.

This paper proposes the evolutionary robust clustering over time (ERCOT) framework to address the above issues. ERCOT is related to the terminology robust optimization over time (ROOT) \cite{fu2015robust} but is significantly different from ROOT in essence. ROOT is for optimization and focuses on predicting robust solutions for future environment. In comparison, ERCOT works on clustering and emphasizes maintaining a robust data partition concerning previous data.

\subsection{Evolutionary Multiobjective Optimization}
Multiobjective optimization problems (MOPs) can be formulated as 
\begin{equation}
	\begin{aligned}
		\min \ \ft(\xt)=&(f_1(\xt),f_2(\xt),\cdots,f_p(\xt)) \\
		                &{\rm s.t.} \ \xt\in\Omega,
	\end{aligned}
\end{equation}
where $\xt=(x_1,x_2,\cdots,x_q)^T$ is a candidate solution, $\Omega\subseteq\mathbb{R}^q$ is the decision space,  $\ft:\Omega\rightarrow\mathbb{R}^p$ includes $p$ conflicting objectives to be minimized, and $\mathbb{R}^p$ is the objective space. EMO algorithms are the main techniques for simultaneously optimizing the multiple objectives in MOPs \cite{liu2019adaptive}\cite{zhao2021evolutionary}.
 
\textbf{Definition 1.} \emph{Solution $\xt_1$ is said to Pareto dominate solution $\xt_2$, i.e., $\xt_1\prec\xt_2$, if and only if $\forall i\in \{1,2,\cdots,p\}, f_i(\xt_1)\leq f_i(\xt_2)$, and $\exists i\in \{1,2,\cdots,p\}, f_i(\xt_1)< f_i(\xt_2)$} \cite{liu2019adaptive}.  

\textbf{Definition 2.} \emph{$\xt^*$ is said to be a Pareto non-dominate solution, if there is no other solution $\xt\in\Omega$ satisfying $\xt\prec\xt^*$} \cite{liu2019adaptive}. 

\textbf{Definition 3.} \emph{All the Pareto non-dominate solutions constitute the Pareto non-dominate solution set $\mathrm{PS}$, the corresponding objective values form the Pareto front (PF)} \cite{liu2019adaptive}.

Clustering can be modeled as MOPs and solved by EMO algorithms. Such clustering paradigm allows simultaneously processing multiple clustering validity indexes and problem/model-specific criteria, offering great flexibility and scalability over traditional methods \cite{mukhopadhyay2015survey}\cite{liu2019transfer}.  

\section{ERCOT}\label{secAlgorithm}
The proposed ERCOT processes the smoothness in temporal data clustering \emph{a posteriori} and automatically tunes the smoothness weight. This new scheme avoids unexpected convergence that occurs in existing methods and holds better scalability.

\subsection{Workflow of ERCOT} 
We take a two-objective function (clustering validity index) and genetic algorithm as an example to illustrate the workflow of ERCOT. However, this illustration does not indicate that ERCOT is tailored to the exemplified objective function and algorithm. Instead, the novelty and core idea of ERCOT, i.e., processing temporal smoothness \emph{a posteriori} and automatically tuning smoothness weight, can be incorporated with various objective functions and algorithms for different problem-solving.  

The workflow is given in Algorithm \ref{algWorkflow}. It starts with randomly generating $N$ solutions (solution representation is explained in Section \ref{sec_encoding}) and evaluating the $N$ solutions' fitness (i.e., clustering validity) by Equation \eqref{eq_obj} (line 2 of Algorithm \ref{algWorkflow}). These solutions and their fitness constitute the initial parent population $\Pt^{G,t}$. Each row of $\Pt^{G,t}$ reports a solution and its corresponding fitness. Iterative generations follow the initialization. Four steps are involved in each generation $G$: 

\begin{algorithm}[t]
	\small  
	\caption{Workflow of ERCOT} \label{algWorkflow}
	\textbf{Require}: maximum number of generations $G_{max}$ at each time step, data $\At^t$ at time $t,t=1,2,\cdots$ 
	\begin{algorithmic}[1]   
		\STATE $G=1, t=1$
		\STATE $\Pt^{G,t} \leftarrow$ initialization($\At^t$)  
		\WHILE{\emph{stopping criterion not met}}
		\IF{\emph{a new time step arrives}}
		\STATE $G=1, t=t+1$
		\STATE $\Pt^{G,t} \leftarrow$ reinitialization($\Pt^{G_{max},t-1},\At^t$)		
		\ENDIF
		\STATE $\Qt^{G,t} \leftarrow$ reproduction($\Pt^{G,t},\At^t$) 
		\IF{$t>1$ and $G=G_{max}$}
		\STATE $\alpha^{t-1}\leftarrow$ smoothness weight auto-tuning($\Pt^{G_{max},t-1},$
		       \\\qquad\qquad $\At^{t-1},\Pt^{G,t},\Qt^{G,t},\At^t$)
		\STATE $\mathring{\Pt}^{G,t},\mathring{\Qt}^{G,t}\leftarrow$ fitness accumulation($\Pt^{G,t},\Qt^{G,t},\At^{t-1},$
		       \\\qquad\qquad\qquad\ $\alpha^{t-1})$
		\STATE $\Pt^{G+1,t}\leftarrow$ environmental selection($\mathring{\Pt}^{G,t},\mathring{\Qt}^{G,t}$)
		\ELSE
		\STATE $\Pt^{G+1,t} \leftarrow$ environmental selection($\Pt^{G,t},\Qt^{G,t}$)
		\ENDIF 
		\STATE $G=G+1$
		\ENDWHILE  
	\end{algorithmic}
	\textbf{Return}: Final partitions of time $t=1,2,\cdots$ 
\end{algorithm}

Step 1, reinitializing the population only if a new time step arrives (lines 4-7 of Algorithm \ref{algWorkflow}). DNSGAII's reinitialization setting \cite{deb2007dynamic} is adopted due to its simplicity and efficiency: $p$ percent of the $N$ solutions are randomly selected from the final population of the last time step $\Pt^{G_{max},t-1}$ to inherit historical partitions (requires a decoding-encoding operation of Equations \eqref{eq_decoding} and \eqref{eq_encoding}), and the other solutions are randomly reinitialized to introduce diversity. These solutions' fitness are evaluated by Equation \eqref{eq_obj}. 

Step 2, reproducing offspring population (line 8 of Algorithm \ref{algWorkflow}). $N$ offspring solutions are produced by the single-point crossover \cite{deb1995simulated} and polynomial mutation \cite{deb1996combined} due to their good performance found in evolutionary clustering literature \cite{mukhopadhyay2015survey}. These solutions are then refined by Equation \eqref{eq_encoding} to accelerate convergence, and their fitness is evaluated by Equation \eqref{eq_obj}. These solutions and their corresponding fitness constitute the offspring population $\Qt^{G,t}$. Each row of $\Qt^{G,t}$ reports a solution and its corresponding fitness.

Step 3, tuning smoothness weight and calculating the accumulated fitness only at the last generation $G_{max}$ of each time $t>1$ (lines 9-12 of Algorithm \ref{algWorkflow}). The weight of smoothness is automatically tuned by quantifying the significance of historical data $\At^{t-1}$ on clustering present data $\At^t$. The accumulated fitness considers not only solutions' accuracy on clustering $\At^t$ but also their quality on partitioning $\At^{t-1}$. Therefore, partition's temporal smoothness will be maintained by selecting solutions according to the accumulated fitness at the last generation (line 12 of Algorithm \ref{algWorkflow}). The accumulated fitness of parent and offspring solutions is updated to $\mathring{\Pt}^{G,t}$ and $\mathring{\Qt}^{G,t}$, respectively. Each row of $\mathring{\Pt}^{G,t}$ and $\mathring{\Qt}^{G,t}$ reports a solution and its corresponding accumulated fitness. 

Step 4, conducting environmental selection. NSGAII's environmental selection \cite{deb2002fast} is utilized to select promising solutions $\Pt^{G+1,t}$ for the next generation (line 12 or 14 of Algorithm \ref{algWorkflow}).

After generation terminates, ERCOT outputs the fittest solution as the final partition. For the two-objective function employed here, the method of \cite{handl2007evolutionary}\cite{garza2018improved} can be used to identify the knee point among the Pareto non-dominate solutions as the final partition. 



\subsection{Objective Function}                                               
We employ a two-objective function. The motivation of employing this objective function is the two objectives trading off each other's tendency to increase or decrease the number of clusters, which enables ERCOT to automatically determine the number of clusters in a single run:
\begin{equation} \label{eq_obj}
    \begin{aligned}
		&\min\ft^t(\xt)=(f_{cp}^t(\zt), f_{sep}(\zt)), \\
		&f_{cp}^t(\zt)=\sum_{c=1}^C \sum_{s=1}^S u_{sc} Dis(\zt_c,\at_s^t), \\
		&f_{sep}(\zt)=1/\min_{c\neq c'}Dis(\zt_c,\zt_{c'}), \\
	\end{aligned}   
\end{equation}
with
\begin{equation*}
	    u_{sc}=\left\{
	\begin{aligned}
		1, \ & if \ Dis(\zt_c,\at_s^t)\leq Dis(\zt_{c'},\at_s^t), \ for \ 1\leq c'\leq C; \\
		0, \ & else, \\
	\end{aligned}\right. 
\end{equation*}
where $\zt$ represents cluster centroids being decoded from solution $\xt$; $\zt_c\in\zt$ ($c=1,2,\cdots,C$) is the $c$th centroid; $\at_s^t\in\At^t$ ($s=1,2,\cdots,S$) is the $s$th data sample at time $t$; and $Dis()$ calculates the Euclidean distance. $f_{cp}$  \cite{handl2007evolutionary} expresses the intra-cluster compactness and is less biased towards spherically shaped clusters compared with the well-known $J_m$ \cite{bezdek1981pattern} which squares $Dis()$. $f_{sep}$ measures the inter-cluster separation \cite{mukhopadhyay2015survey}.

The advantages of the two-objective function over existing ones in temporal data clustering are that it is conceptually simple and captures more diverse characteristics of datasets by jointly considering the intra- and inter-cluster information. More importantly, $f_{cp}$ prefers partitions with a large number of clusters, whereas $f_{sep}$ favors partitions with fewer clusters, i.e., the two objectives balance each other's tendency to increase or decrease the number of clusters. Consequently, multiple partitions at different hierarchical levels can be obtained in a single algorithmic run. The final number of clusters can further be determined by selecting a final partition from the obtained ones. 

Note that apart from the cluster prototype-based $f_{cp}$ and $f_{sep}$, the validity indexes based on cluster labels can also be incorporated with ERCOT for distinctively shaped datasets \cite{mukhopadhyay2015survey}.

\subsection{Solution Encoding}\label{sec_encoding}
\begin{figure}[t] 
	\centering
	\includegraphics[width=1\linewidth]{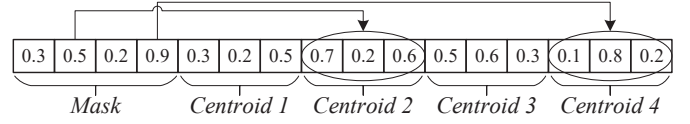}
	\caption{An example of solution encoding with $C_{max}=4$ and $D=3$, where $C_{max}$ is the maximum number of clusters, and $D$ is the dimensionality of data samples.}
	\label{figencoding}
\end{figure}

We adopt a hybrid real encoding to fit with the employed objective function, as shown in Figure \ref{figencoding}. The motivation of adopting this encoding is it allowing ERCOT to automatically optimize the number of clusters $C$ ($2\leq C\leq C_{max}$). The first $C_{max}$ entries are masks controlling the activation/inactivation of cluster centroids, while the rest entries represent cluster centroids. The mask $m_c$ ($c=1,2,\cdots,C_{max}$) is a real number from $[0,1]$. Centroid $\zt_c$ is activated if $m_c\geq0.5$; otherwise $\zt_c$ is inactivated. The example in Figure \ref{figencoding} represents a partition with $C=2$ clusters, and the centroids are $(0.7,0.2,0.6)$ and $(0.1,0.8,0.2)$, respectively. 

Note that the Locus-based encoding can also be adopted if cluster label-based validity indexes are involved in the objective function for distinctively shaped datasets \cite{hancer2017comprehensive}. 

\textbf{Remark 1}. Since the data evolve over time, the centroids represented by solutions in $\Pt^{G_{max},t-1}$ cannot be directly used for the reinitialization of time $t$ (line 6 of Algorithm \ref{algWorkflow}). We conduct a decoding-encoding operation to transform the centroids from the data space of time $t-1$ to that of time $t$ as follows.

For each of the selected solutions from $\Pt^{G_{max},t-1}$, we decode its activated centroids $\zt_c$, $c=1,2,\cdots,C$, to the memberships of data samples:
\begin{equation}\label{eq_decoding}
	u_{sc}=\left\{
	\begin{aligned}
		1, \ & if \ Dis(\zt_c,\at_s^{t-1})\leq Dis(\zt_{c'},\at_s^{t-1}), \ for \ 1\leq c'\leq C; \\
		0, \ & else, \\
	\end{aligned}\right. 
\end{equation}
where $u_{sc}$ is the membership of the sample $s$ to centroid $c$ at time $t-1$. After that, we update the activated centroids by encoding the memberships according to $\At^t$:
\begin{equation}\label{eq_encoding}
	\zt_c=\sum_{s=1}^Su_{sc}\at_s^{t} \left/ \sum_{s=1}^Su_{sc}\right..	
\end{equation}
Only the samples that appear at both time $t$ and $t-1$ are considered in Equation \eqref{eq_encoding}. Solutions with updated activated centroids (masks and inactivated centroids remain unchanged) inherit time $(t-1)$'s partitions and fit with time $t$'s data prototype, thus can be used as initial solutions of $t$.    

\textbf{Remark 2}. To accelerate convergence, at the reproduction step (line 8 of Algorithm \ref{algWorkflow}), we refine the activated centroids $\zt_c$, $c=1,2,\cdots,C$, of each offspring solution by Equation \eqref{eq_encoding}, where $u_{sc}=1$, if $Dis(\zt_c,\at_s^t)\leq Dis(\zt_{c'},\at_s^t)$, for $1\leq c'\leq C$; $u_{sc}=0$, otherwise.

\subsection{Fitness Accumulation}
We propose the fitness accumulation to measure solutions' clustering validity on both the present data $\At^t$ and the previous data $\At^{t-1}$ at the last generation $G_{max}$ of each time $t>1$. The temporal smoothness of partition can be maintained by selecting solutions according to the accumulated fitness (line 12 of Algorithm \ref{algWorkflow}). We define the accumulated fitness $\mathring{\ft}^t(\xt)$ of solution $\xt$ at time $t$ ($t=2,3,\cdots$) as
\begin{equation} \label{eq_accumulatedobj}
	\mathring{\ft}^t(\xt)=(1-\alpha^{t-1})\ft^t(\xt)+\alpha^{t-1}\ft^{t-1}(\xt),   
\end{equation}
where $\ft^t(\xt)$ and $\ft^{t-1}(\xt)$ are calculated by Equation \eqref{eq_obj}, $\ft^t(\xt)$ is the fitness of $\xt$ on the present data $\At^t$, $\ft^{t-1}(\xt)$ is the fitness of $\xt$ on the previous data $\At^{t-1}$, $\alpha^{t-1}$ is the weight of temporal smoothness and is automatically tuned by the method in Section \ref{sec_weight}. Since the data evolve over time, before evaluating $\ft^{t-1}(\xt)$, the decoding-encoding operation of Equations \eqref{eq_decoding} and \eqref{eq_encoding} is executed to transform the centroids represented by $\xt$ from the data space of time $t$ to that of $t-1$. Parents and offspring with accumulated fitness are updated to $\mathring{\Pt}^{G,t}$ and $\mathring{\Qt}^{G,t}$, respectively (line 11 of Algorithm \ref{algWorkflow}).

The proposed fitness accumulation has two key features: 1) It expresses the temporal smoothness of how well the current partition fits with historical data, avoiding being affected by historical partition error. 2) It is executed only at the last generation of each time step, enabling ERCOT to search for accurate partitions without being penalized by temporal smoothness (lines 8 and 14 of Algorithm \ref{algWorkflow}), and then process the smoothness in an \emph{a posterior} manner (lines 9-12 of Algorithm \ref{algWorkflow}). 

This \emph{posterior} manner is rational due to the following reasons. First, it is intuitive to prioritize clustering accuracy then process smoothness \emph{a posterior}, as a smooth partition is insignificant without recognizing the actual data structure. Second, by inheriting historical information in the reinitialization step (line 6 of Algorithm \ref{algWorkflow}), the partitions obtained currently would naturally not deviate too much from the history provided that the data at adjacent time periods are similar. Third, this manner avoids the potential unexpected convergence caused by improper \emph{prior} preference in existing algorithms.  

\subsection{Smoothness Weight Auto-Tuning}\label{sec_weight}
We propose to automatically tune the weight of temporal smoothness $\alpha^{t-1}$ by quantifying the significance of historical data $\At^{t-1}$ on clustering present data $\At^t$. The idea of quantification is inspired by the stacked density estimation \cite{smyth1997stacked}\cite{da2019curbing}. In detail, we stack the probabilistic model of $\At^t$ with that of $\At^{t-1}$; then, we find the optimal stack which maximally approximates the true latent probability density function describing $\At^t$. The weight of $\At^{t-1}$ in the optimal stack represents the significance of $\At^{t-1}$. As a clusterable dataset has multiple clusters with different statistical characteristics, the quantification is conducted in each cluster rather than in the whole dataset. 
 
\begin{figure*}[tb] 
	\centering
	\includegraphics[width=1\linewidth]{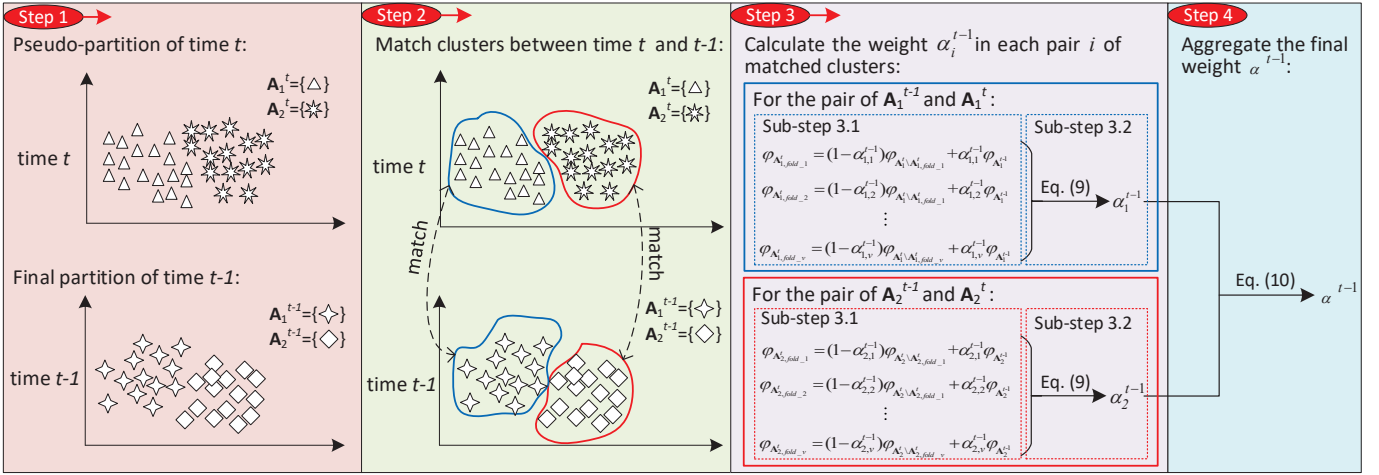}
	\caption{Procedure of smoothness weight auto-tuning at time $t$.}
	\label{fig_weight}
\end{figure*}

The procedure of smoothness weight auto-tuning is depicted in Figure \ref{fig_weight}. In this procedure, first, $\At^t$ and $\At^{t-1}$ are normalized to zero mean, respectively; then, only the data samples of $\At^t$ and $\At^{t-1}$ that appear at both time $t$ and $t-1$ are considered in the following four steps. Since only the common data samples at adjacent time steps are needed in processing smoothness, ERCOT can readily handle datasets with inserting and removing samples over time.   

In step 1, to conduct the quantification in each cluster, we need a pseudo-partition of $\At^t$ and link the clusters formed by the pseudo-partition to $\At^{t-1}$'s clusters\footnote{$\At^{t-1}$'s clusters have already been obtained at time $t-1$.}. As the smoothness weight auto-tuning is executed \emph{a posterior} after data clustering (line 9 of Algorithm \ref{algWorkflow}), we can identify the knee point among the Pareto non-dominated solutions of the final population $\Pt^{G_{max},t}\cup\Qt^{G_{max},t}$ as the pseudo partition of $\At^t$ by the method of \cite{handl2007evolutionary}. This pseudo partition is a feasible classification of $\At^t$ without penalizing by temporal smoothness.

In step 2, we execute a cluster matching operation since clusters' permutation at time $t$ may be different from that at $t-1$. For each cluster in the pseudo-partition of time $t$, this operation pairs it with a $(t-1)$'s cluster, such that the two paired clusters have the maximum number of common samples. This one-to-one matching can be done in polynomial time by the Hungarian algorithm \cite{kuhn1955hungarian}. 

In step 3, for each pair of the matched clusters, we stack the probabilistic models as:
\begin{equation}\label{eq_stack}
	\tilde{\varphi}_{\At_i^t}=(1-\alpha_i^{t-1})\varphi_{\At_i^t}+\alpha_i^{t-1}\varphi_{\At_i^{t-1}},
\end{equation} 
where $i$ refers to the $i$th pair of matched clusters; ${\At_i^{t-1}}$ and $\At_i^t$ are the data belonging to the matched clusters at time $t-1$ and $t$, respectively; $\varphi_{\At_i^{t-1}}$ and $\varphi_{\At_i^t}$ are the probabilistic models of ${\At_i^{t-1}}$ and $\At_i^t$, respectively; $\tilde{\varphi}_{\At_i^t}$ is an approximation of $\bar{\varphi}_{\At_i^t}$, where $\bar{\varphi}_{\At_i^t}$ is the true latent probability density function describing $\At_i^t$; $\alpha_i^{t-1}$ represents the significance of $\At_i^{t-1}$.  

The optimal stack can be obtained by maximizing the probability of observing out-of-sample $\At_i^t$, which can be modeled as maximizing the log-likelihood function \cite{da2019curbing}:    
\begin{equation}\label{eq_likelihood}
    \log L = \sum_{\at\in\At_{test}}\log\tilde{\varphi}_{\At_i^t},
\end{equation}
where $\At_{test}$ are the out-of-sample (test) data of $\At_i^t$. Equation \eqref{eq_likelihood} can be solved by the method of \cite{da2019curbing} in two sub-steps:
\begin{itemize} 
	\item Sub-step 3.1, randomly dividing $\At_i^t$ into $v$ folds by the principle of cross-validation. For each fold, the \emph{training} part is used to deduce the $\varphi_{\At_i^t}$ of Equation \eqref{eq_stack}; the \emph{testing} part forms the $\At_{test}$ of Equation \eqref{eq_likelihood}. The likelihood of each sample in the \emph{testing} part can be estimated by Equations \eqref{eq_stack} and \eqref{eq_likelihood}.
	
	By repeating the above procedure in all folds, a $|\At_i^t|\times2$ matrix is obtained, where $|\At_i^t|$ is the number of samples in $\At_i^t$. The $(s,1)$th and $(s,2)$th entries of the matrix are the out-of-sample likelihood of $\varphi_{\At_i^t}$ on $\at_s$ and that of $\varphi_{\At_i^{t-1}}$ on $\at_s$, respectively, where $\at_s$ is the $s$th sample of $\At_i^t$. 
	\item Sub-step 3.2, estimating the $\alpha_i^{t-1}$ of the optimal stack by maximizing the equivalence of Equation \eqref{eq_likelihood}:
	\begin{equation}\label{eq_likelihood2}
		\log L = \sum_s^{|\At_i^t|}\log((1-\alpha_i^{t-1})\varphi_{\At_i^t}(\at_j)+\alpha_i^{t-1}\varphi_{\At_i^{t-1}}(\at_j)),
	\end{equation}
	where $\varphi_{\At_i^t}(\at_s)$ and $\varphi_{\At_i^{t-1}}(\at_s)$ are the $(s,1)$th and $(s,2)$th entries of the matrix obtained in sub-step 3.1, respectively. Equation \eqref{eq_likelihood2} can be easily solved by the expectation-maximization algorithm \cite{moon1996expectation}.  
\end{itemize}
      
In step 4, after estimating the weights of all pairs of matched clusters, we aggregate the final weight $\alpha^{t-1}$ according to the size of each cluster:
\begin{equation}
	\alpha^{t-1}=\sum_i^{C^*}\frac{|\At_i^{t-1}|+|\At_i^t|}{|\At^{t-1}|+|\At^t|}\alpha_i^{t-1},
\end{equation}   
where $|\At|$ denotes the number of samples in $\At$, and $i=1,2,\cdots,C^*$ refers to the $i$th pair of matched clusters. A higher $\alpha^{t-1}$ value indicates a larger relevance of $\At^{t-1}$ to $\At^t$, and vice versa.   

\subsection{Computational Complexity}
The computational complexity of each generation of ERCOT is analyzed below. 

The two objectives of Equation \eqref{eq_obj}, i.e., $f_{cp}$ and $f_{sep}$ require $\mathcal{O}(SDC_{max}N)$ and $\mathcal{O}(DC_{max}(C_{max}-1)N)$ computations, respectively, where $S$ is the number of data samples to be clustered, $D$ is the dimensionality of data sample, $C_{max}$ is the maximum number of clusters, and $N$ is the population size. Because $C_{max}-1\ll S$, the computational complexity of objective function evaluation is $\mathcal{O}(SDC_{max}N)$. 

The complexity of evolutionary operators, i.e., the crossover, mutation, and selection, is dominated by environmental selection \cite{deb2002fast}. The complexity of environmental selection is $\mathcal{O}(pN^2)$ \cite{deb2002fast}, where $p$ is the number of objectives. ERCOT has $p=2$ objectives, thus the complexity of environmental selection is $\mathcal{O}(N^2)$. 

The complexity of smoothness weight auto-tuning is $\mathcal{O}(SDv)$, where $v$ is the number of folds of the cross-validation employed in Sub-step 3.1. 

Because $N\ll S$, the overall complexity of ERCOT is $\mathcal{O}(SDC_{max}N+SDv)$. 

\section{Experiment}\label{secExp}
\subsection{Experimental Setup}
\subsubsection{Datasets}
We design eight synthetic datasets emulating various kinds of change patterns, i.e., data evolve with 1) varying attribute values, 2) concept drift \cite{chi2009evolutionary}, 3) inserting and removing samples, and 4) varying cluster numbers. These datasets comprehensively verify the efficiency and effectiveness of ERCOT. Details of the datasets are given along with the experimental analysis in Section \ref{sec_syn_exp}. We further examine ERCOT on two real (experimental) temporal data clustering problems, i.e., bird flocks tracking \cite{reynolds1987flocks} and video background segmentation \cite{chacon2018bio}, to testify ERCOT's practicability.

\subsubsection{Measures}
We employ the rand index (RI) \cite{rand1971objective} and NMI \cite{strehl2002cluster} as performance measures. RI evaluates the similarity between the clustering result and data labels, whereas NMI calculates the mutual information between the clustering result and data labels. Higher RI and NMI values indicate better performance. For temporal data, we calculate the mean RI (mRI) and mean NMI (mNMI) of all time steps to evaluate algorithms' overall performance.

\subsubsection{Referred Algorithms}
We compare ERCOT with five temporal data clustering algorithms\footnote{There are also relevant algorithms \cite{li2015sparsity}\cite{wang2017hierarchical}\cite{hashemi2018evolutionary}\cite{arzeno2017evolutionary}\cite{arzeno2019evolutionary}, but they are specific to co-clustering \cite{li2015sparsity}, text data clustering \cite{wang2017hierarchical}, subspace clustering \cite{hashemi2018evolutionary}, and offline clustering (both historical and future data are used for temporal smoothness) \cite{arzeno2017evolutionary}\cite{arzeno2019evolutionary}, thus are unavailable for comparison.}, i.e., \emph{k}-meansCOT \cite{chakrabarti2006evolutionary}, AFFECT \cite{xu2014adaptive}, E$^2$SC \cite{langone2016efficient}, penaltyEC, and staticEC:

\begin{itemize} 
	\item \emph{k}-meansCOT \cite{chakrabarti2006evolutionary}: A seminal algorithm which incorporates temporal smoothness into the centroid update step of \emph{k}-means.
	\item AFFECT \cite{xu2014adaptive}: A state-of-the-art method that adaptively estimates the weight of smoothness. Here we use the \emph{k}-means-based AFFECT version. 
	\item E$^2$SC \cite{langone2016efficient}: A state-of-the-art spectral clustering approach with a predefined smoothness weight=0.1.
	\item penaltyEC: A variation of ERCOT that we give here for experimental comparison. It employs the temporal data clustering idea in dynamic community detection \cite{folino2014evolutionary}\cite{bara2016new} (mentioned in Section \ref{sec_clustering_review}), since the problem-specific methods of \cite{folino2014evolutionary}\cite{bara2016new} cannot be directly employed for comparison. PenaltyEC uses the same reinitialization, reproduction, and environmental selection with ERCOT but simultaneously optimizes the $f_{cp}$ and the NMI between partitions at adjacent time steps without fitness accumulation. 
	\item staticEC: The static version of ERCOT that clusters the data at each time independently.
\end{itemize}

\subsubsection{Parameter Settings}
Algorithms' parameters are set in accordance with their original papers \cite{chakrabarti2006evolutionary}\cite{xu2014adaptive}\cite{langone2016efficient}, respectively. Parameters in the evolutionary operators in penaltyEC, staticEC, and ERCOT are set following the practice in \cite{deb2007dynamic}\cite{deb1995simulated}, i.e., crossover rate=0.9, mutation rate=0.9, mutation distribution index=20, reinitialization percent $p$=0.8, and population size $N$=100. In the smoothness weight auto-tuning of ERCOT, multi-variate Gaussian distribution is used as the probabilistic model, and $v$=500-fold cross-validation\footnote{The leave-one-out cross-validation is adopted to cases with less than 500 data samples.} is employed for a trade-off between estimation accuracy and time efficiency. The maximum number of clusters $C_{max}=8$ for experiments without giving the number of clusters a priori. The maximum fitness (function) evaluations of each algorithm at each time step is $1000$ for a fair comparison. Each algorithm runs $30$ times on each dataset. 

\subsection{Performance Comparison}\label{sec_syn_exp}
\begin{figure}[t] 
	\centering
	\includegraphics[width=0.9\linewidth]{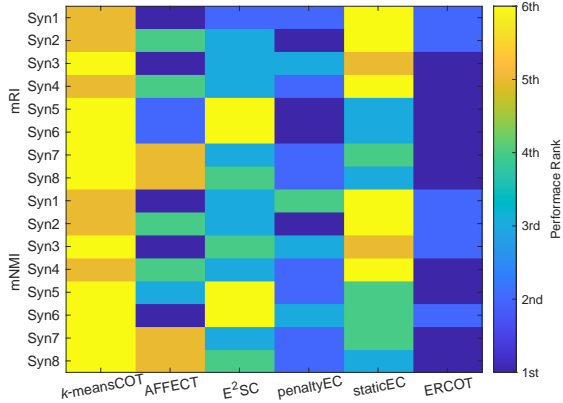}
	\caption{Heat map of each algorithm ranked at a particular position in mRI and mNMI performance comparison in each dataset.}
	\label{figHeatmap}
\end{figure}

The mRI and mNMI results of algorithms on synthetic datasets are collected in Tables \ref{tabRI} and \ref{tabNMI}, respectively. Algorithms' performance ranks are depicted in Figure \ref{figHeatmap}. Figure \ref{figHeatmap} clearly shows that ERCOT wins the best in most datasets, demonstrating ERCOT's overall superiority. Detailed analysis is given below.   

\begin{table*}[t]
	\caption{Mean and standard deviation of mRI results in synthetic datasets. Best values are bold.}
	\centering
	\label{tabRI}
	\begin{threeparttable}
	\begin{tabular}{lllllll}
		\hline
		& \multicolumn{1}{c}{\emph{k}-meansCOT} & \multicolumn{1}{c}{AFFECT} & \multicolumn{1}{c}{E$^2$SC} & \multicolumn{1}{c}{penaltyEC} & \multicolumn{1}{c}{staticEC} & \multicolumn{1}{c}{ERCOT}  \\ \hline
		Syn1 & 0.8988$\pm$1.97E-02$>$ & \textbf{0.9558}$\pm$5.69E-02$\approx$ & 0.9544$\pm$4.52E-16$\approx$ & 0.9545$\pm$1.05E-03$\approx$ & 0.8956$\pm$1.38E-02$>$ & 0.9545$\pm$3.20E-03 \\
		Syn2 & 0.9415$\pm$2.34E-02$>$ & 0.9692$\pm$5.69E-02$>$ & 0.9872$\pm$4.52E-16$>$ & \textbf{0.9927}$\pm$6.71E-04 & 0.8422$\pm$1.76E-02$>$ & 0.9907$\pm$8.70E-04 \\
		Syn3 & 0.9059$\pm$1.51E-02$>$ & \textbf{0.9557}$\pm$5.47E-02$\approx$ & 0.9516$\pm$4.52E-16$>$ & 0.9512$\pm$6.63E-03$>$ & 0.9153$\pm$1.55E-02$>$ & 0.9554$\pm$4.51E-03 \\
		Syn4 & 0.9336$\pm$2.09E-02$>$ & 0.9684$\pm$4.62E-02$>$ & 0.9854$\pm$1.13E-16$>$ & 0.9891$\pm$7.95E-03$>$  & 0.8679$\pm$1.48E-02$>$ & \textbf{0.9949}$\pm$6.33E-04 \\
		Syn5 & n/a & 0.9631$\pm$5.53E-02$>$ & n/a & 0.9791$\pm$9.84E-04$\approx$ & 0.8666$\pm$1.74E-02$>$ & \textbf{0.9792}$\pm$1.01E-03 \\
		Syn6 & n/a & 0.9695$\pm$4.30E-02$>$ & n/a & 0.9714$\pm$1.04E-03$>$ & 0.8958$\pm$1.83E-02$>$ & \textbf{0.9739}$\pm$1.50E-03 \\
		Syn7 & n/a & 0.7230$\pm$1.48E-03$>$ & 0.9489$\pm$6.78E-16$>$ & 0.9501$\pm$4.18E-03$>$ & 0.8877$\pm$1.05E-02$>$ & \textbf{0.9526}$\pm$4.32E-03 \\
		Syn8 & n/a & 0.7313$\pm$1.90E-03$>$ & 0.8570$\pm$3.39E-16$>$ & 0.9768$\pm$7.66E-03$>$ & 0.9207$\pm$1.10E-02$>$ & \textbf{0.9862}$\pm$1.07E-03 \\ \hline
	\end{tabular}
\begin{tablenotes}
	\scriptsize
	\item $>$: The performance of ERCOT is significantly better than the outcome according to the Wilcoxon Sign Test \cite{d1986principles} at the 0.05 significance level.
	\item $\approx$: The performance of ERCOT is comparable with the outcome according to the Wilcoxon Sign Test at the 0.05 significance level.
\end{tablenotes}
\end{threeparttable}
\end{table*}

\begin{table*}[t]
	\caption{Mean and standard deviation of mNMI results in synthetic datasets. Best values are bold.}
	\centering
	\label{tabNMI}
	\begin{threeparttable}
		\begin{tabular}{lllllll}
		\hline
		& \multicolumn{1}{c}{\emph{k}-meansCOT} & \multicolumn{1}{c}{AFFECT} & \multicolumn{1}{c}{E$^2$SC} & \multicolumn{1}{c}{penaltyEC} & \multicolumn{1}{c}{staticEC} & \multicolumn{1}{c}{ERCOT}  \\ \hline
		Syn1 & 0.7697$\pm$3.03E-02$>$ & \textbf{0.9264}$\pm$6.93E-02 & 0.8673$\pm$3.39E-16$>$ & 0.8651$\pm$2.91E-03$>$ & 0.7668$\pm$2.07E-02$>$ & 0.8766$\pm$9.39E-03 \\
		Syn2 & 0.8862$\pm$3.73E-02$>$ & 0.9636$\pm$6.79E-02$\approx$ & 0.9640$\pm$4.52E-16$>$ & \textbf{0.9778}$\pm$2.01E-03 & 0.7438$\pm$3.07E-02$>$ & 0.9693$\pm$7.28E-03 \\
		Syn3 & 0.7786$\pm$2.39E-02$>$ & \textbf{0.9158}$\pm$7.65E-02 & 0.8635$\pm$3.39E-16$>$ & 0.8719$\pm$2.48E-03$>$ & 0.7918$\pm$2.73E-02$>$ & 0.8847$\pm$5.95E-03 \\
		Syn4 & 0.8668$\pm$3.36E-02$>$ & 0.9463$\pm$6.31E-02$>$ & 0.9578$\pm$4.52E-16$>$ & 0.9817$\pm$2.19E-03$>$ & 0.7744$\pm$2.63E-02$>$ & \textbf{0.9839}$\pm$5.45E-03 \\
		Syn5 & n/a & 0.9308$\pm$6.85E-02$>$ & n/a & 0.9330$\pm$2.90E-03$>$ & 0.7583$\pm$2.32E-02$>$ & \textbf{0.9381}$\pm$2.23E-03 \\
		Syn6 & n/a & \textbf{0.9368}$\pm$5.80E-02 & n/a & 0.9194$\pm$2.58E-03$>$ & 0.7969$\pm$2.70E-02$>$ & 0.9267$\pm$2.22E-03 \\
		Syn7 & n/a & 0.6239$\pm$1.92E-02$>$ & 0.8056$\pm$4.52E-16$>$ & 0.8296$\pm$1.35E-02$>$ & 0.7580$\pm$1.55E-02$>$ & \textbf{0.8576}$\pm$5.85E-03 \\
		Syn8 & n/a & 0.6469$\pm$4.76E-03$>$ & 0.7527$\pm$0.00E+00$>$ & 0.9398$\pm$1.37E-02$>$ & 0.8529$\pm$1.64E-02$>$ & \textbf{0.9616}$\pm$3.16E-03 \\ \hline
		\end{tabular}
		\begin{tablenotes}
			\scriptsize
			\item $>$: The performance of ERCOT is significantly better than the outcome according to the Wilcoxon Sign Test at the 0.05 significance level.
			\item $\approx$: The performance of ERCOT is comparable with the outcome according to the Wilcoxon Sign Test at the 0.05 significance level.
		\end{tablenotes}
	\end{threeparttable}
\end{table*}

\subsubsection{Tracking Evolving Clusters}
\begin{figure}[tb] 
	\centering
	\includegraphics[width=0.9\linewidth]{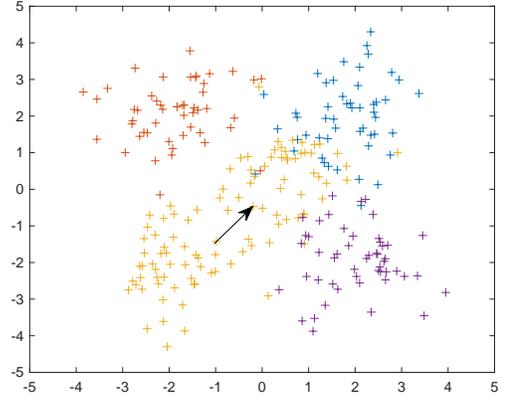}
	\caption{Illustration of synthetic dataset in the 2-D space. The bottom-left cluster moves toward the top-right one as time goes by.}
	\label{figsyndata}
\end{figure}

Syn1 and Syn2 test the algorithms' ability to track evolving clusters. We design the two datasets similar to those of \cite{xu2014adaptive}\cite{langone2016efficient}. For Syn1, 200 samples are drew equally from four 2-D Gaussian distributions with mean vectors $(2,2)$, $(-2,2)$, $(-2,-2)$, $(2,-2)$ and covariance matrices $0.8I$ ($I$ is the identity matrix), respectively, as shown in Figure \ref{figsyndata}. We consider a horizon with 10 time steps and move the bottom-left cluster towards the top-right one with step size $0.6$ at each time. The clusters overlap during time 4 to 7, which poses difficulty in tracking the evolving clusters. Syn2 is a 4-D version of Syn1 with mean vectors $(2,2,2,2)$, $(-2,-2,2,2)$, $(-2,-2,-2,-2)$, and $(2,2,-2,-2)$. Other settings of Syn2 are the same as Syn1. We do not consider higher-dimensional instances because handling high-dimensional data is out of the scope of this paper.  

According to Tables \ref{tabRI} and \ref{tabNMI}, the proposed ERCOT is overall comparable with AFFECT, E$^2$SC, and penaltyEC, and is significantly better than \emph{k}-meansCOT and staticEC in Syn1. The four leading algorithms' performance further enhances in Syn2 because data in 4-D space are more distinguishable. These results demonstrate that ERCOT can efficiently track the evolving clusters by inheriting historical partitions (line 6 of Algorithm \ref{algWorkflow}) and considering temporal smoothness via \emph{a posterior} fitness accumulation (line 11 of Algorithm \ref{algWorkflow}). By contrast, staticEC fails to handle the two datasets without historical information's assistance to partition overlapped clusters.      

\subsubsection{Handling Concept Drift}
Syn3 and Syn4 examine algorithms' performance on handling concept drift. Syn3 is a variation of Syn1, which has the following difference from Syn1. In time 4, 15 samples of the bottom-left cluster (refer to Figure \ref{figsyndata}) switch their memberships (i.e., concept) to the other three clusters due to cluster overlapping. In time 5, another 15 samples of the bottom-left cluster move to the other three clusters. In time 6, the 15 samples mentioned at time 5 move back to the bottom-left cluster. Syn4 is a variation of Syn2, which has the same membership switch as the above.   

\begin{figure}[t] 
	\centering
	\includegraphics[width=0.9\linewidth]{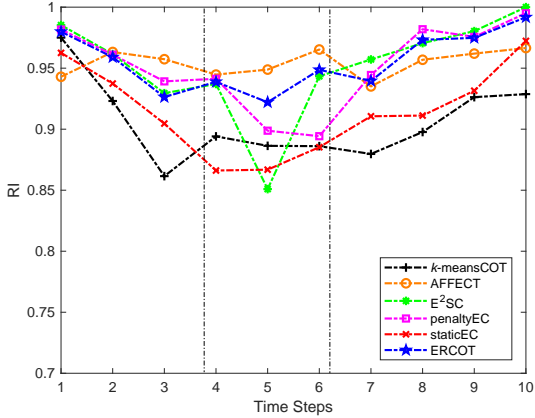}
	\caption{Average RI value among all runs of each algorithm at each time step of Syn3.}
	\label{figRIsyn3}
\end{figure}

From Tables \ref{tabRI} and \ref{tabNMI}, ERCOT wins the best in both Syn3 and Syn4, illustrating its outperformance in handling concept drift. The RI results versus time steps of Syn3 are given in Figure \ref{figRIsyn3}. According to Figure \ref{figRIsyn3}, ERCOT and AFFECT achieve promising and stable results over all time periods. Results of E$^2$SC fluctuate dramatically during time 4 to 6, indicating that the fixed weight of temporal smoothness in E$^2$SC is not suitable when sudden change (i.e., the concept drift) happens. The performance of penaltyEC deteriorates from time 4 to 6. The reason is possible that penaltyEC maintains temporal smoothness by maximizing the consistency of data samples' memberships between adjacent time steps, which is improper in concept drift (i.e., membership switch) scenarios.  
          
\subsubsection{Handling Time-Varying Samples}
Syn5 and Syn6 investigate algorithms' performance on cases with inserting and removing samples over time. Syn5 is a variation of Syn1, which has the following difference with Syn1. In time 4, 10 samples are randomly selected and removed from each cluster. In time 6, 10 new samples are drew for  each cluster. In time 8, 10 samples are randomly selected and removed from each cluster. Syn6 is a variation of Syn3, which has the same sample change as the above.   

The proposed ERCOT is in leading position (from Tables \ref{tabRI} and \ref{tabNMI}) in both Syn5 and Syn6, because it can readily handle such datasets by only considering the common samples in the interaction of adjacent time steps (Lines 6, 10, 11 of Algorithm \ref{algWorkflow}). The penaltyEC is competitive in Syn5 but is inferiors to ERCOT in Syn6. The reason is similar with that in Syn3. \emph{k}-meansCOT and E$^2$SC cannot process datasets with time-varying samples, hence having no scores in Syn5 and Syn6. 


\subsubsection{Handling Time-Varying Clusters and Determining Cluster Numbers}\label{sec_num_cluster}
Syn7 and Syn8 exploit algorithms' ability to handle time-varying clusters and determine the number of clusters. Syn7 is derived from Syn3, in which the bottom-left cluster disappears at times 5 and 6, i.e., samples of the bottom-left cluster move to other clusters at the two time steps. Syn8 is derived from Syn4 in the same way as the above. Syn7 and Syn8 pose great challenges to algorithms because they are characterized as both time-varying clusters and concept drift. The actual number of clusters is set to be unknown beforehand in the experiment on these two datasets.  

It can be observed from Tables \ref{tabRI} and \ref{tabNMI} that ERCOT constantly obtains better mRI and mNMI results than other algorithms in Syn7 and Syn8. This superiority benefits from the proposed two-objective function (Equation \eqref{eq_obj}) and the employed final solution identification method \cite{handl2007evolutionary}. PenaltyEC is worse than ERCOT in Syn7 and Syn8 because its objectives cannot balance each other's tendency to increase or decrease cluster numbers, thereby failing to determine a proper number of clusters. The performance of AFFECT is unacceptable, indicating that the Silhouette index \cite{rousseeuw1987silhouettes} that it employed to determine the number of clusters does not work as expected.

\subsection{Efficiency of Core Components of ERCOT}
\subsubsection{Efficiency of Objective Function} 
\begin{figure}[tb] 
	\centering
	\includegraphics[width=1\linewidth]{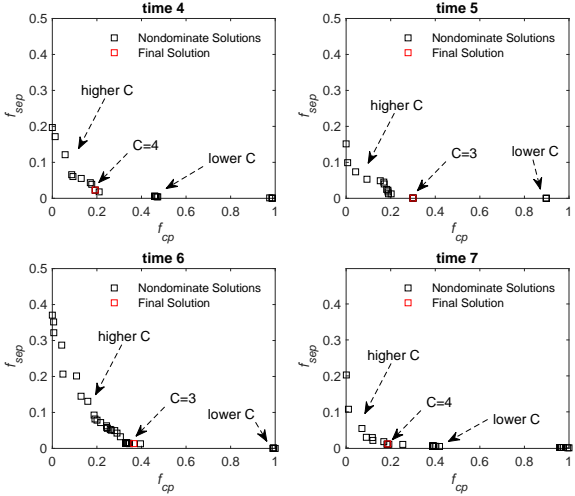}
	\caption{Solutions of ERCOT to time 4-7 of Syn7 depicted in the normalized objective space. C is the number of clusters.}
	\label{figPF}
\end{figure}

The two proposed objectives are conflicting in terms of the tendency to the number of clusters. This characteristic enables ERCOT to determine the number of clusters automatically, and the efficiency has been verified in Section \ref{sec_num_cluster}. Figure \ref{figPF} further plots the solutions obtained by ERCOT at a single run in Syn7. This figure shows that the two objectives hold a trade-off in the number of clusters. A range of solutions with different numbers of clusters are obtained, and the final number of clusters is automatically determined by selecting a final solution from the obtained ones. These results demonstrate the efficiency of the objective function of ERCOT.  

\subsubsection{Efficiency of Fitness Accumulation} The fitness accumulation expresses temporal smoothness and is only executed at the last generation of each time step. This setting enables ERCOT to search for accurate partitions without being penalized by temporal smoothness, and then process the smoothness in an \emph{a posterior} manner. The validity of fitness accumulation can be testified by comparing it with penaltyEC which uses the same evolutionary operators as ERCOT but processes the smoothness in an \emph{a priori} manner. According to Tables \ref{tabRI} and \ref{tabNMI}, among the 12 mRI and mNMI comparisons in Syn1-6, ERCOT outperforms penaltyEC in 8 instances and is comparable with penaltyEC in 2 out of rest 6 cases. These results demonstrate the efficiency of \emph{a posterior} fitness accumulation. 

\subsubsection{Efficiency of Smoothness Weight Auto-Tuning} 
\begin{figure}[t]
	\centering
	\subfigure[\label{fig_w_syn3}]{\includegraphics[width=1.7in]{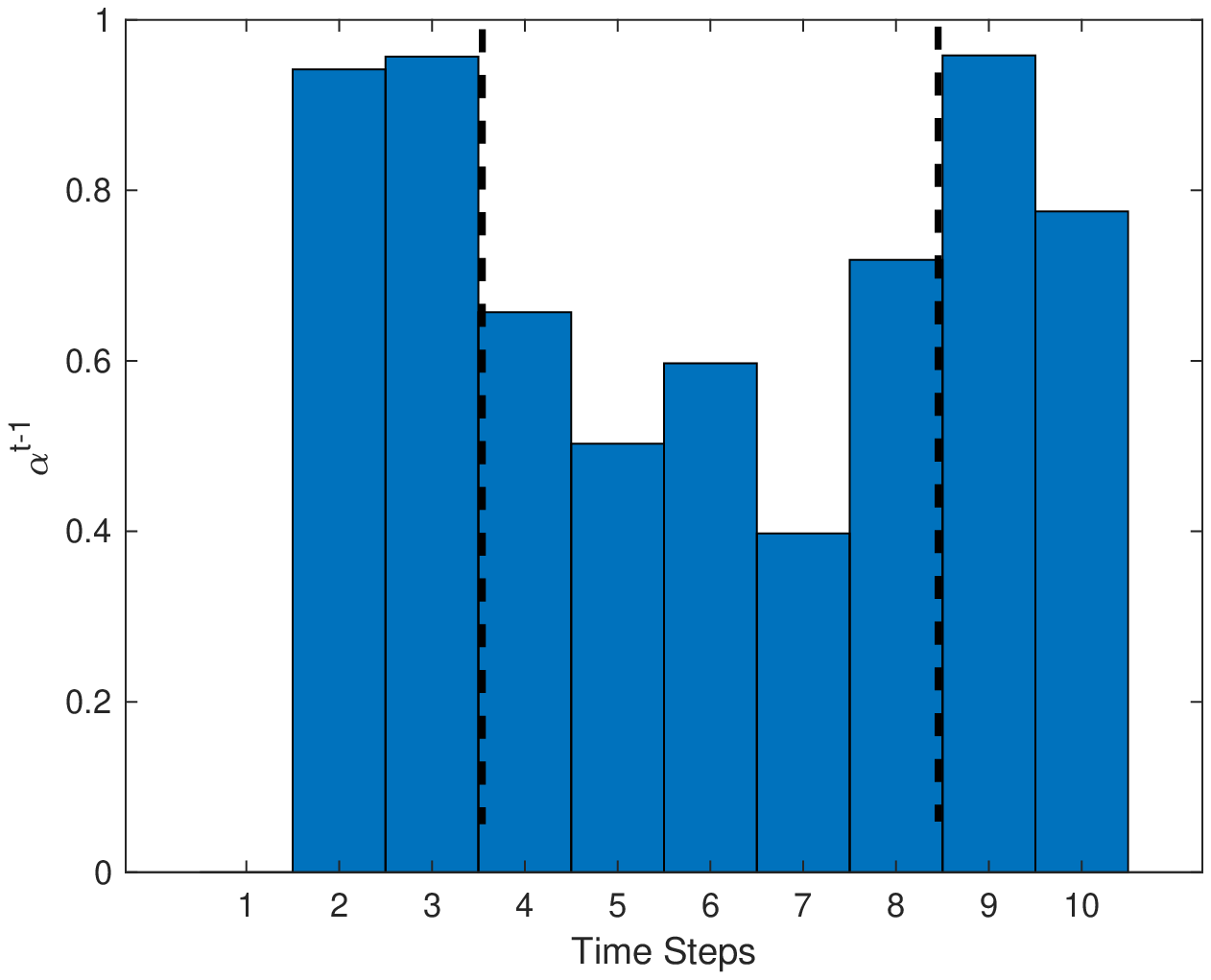}}
	\subfigure[\label{fig_w_syn7}]{\includegraphics[width=1.7in]{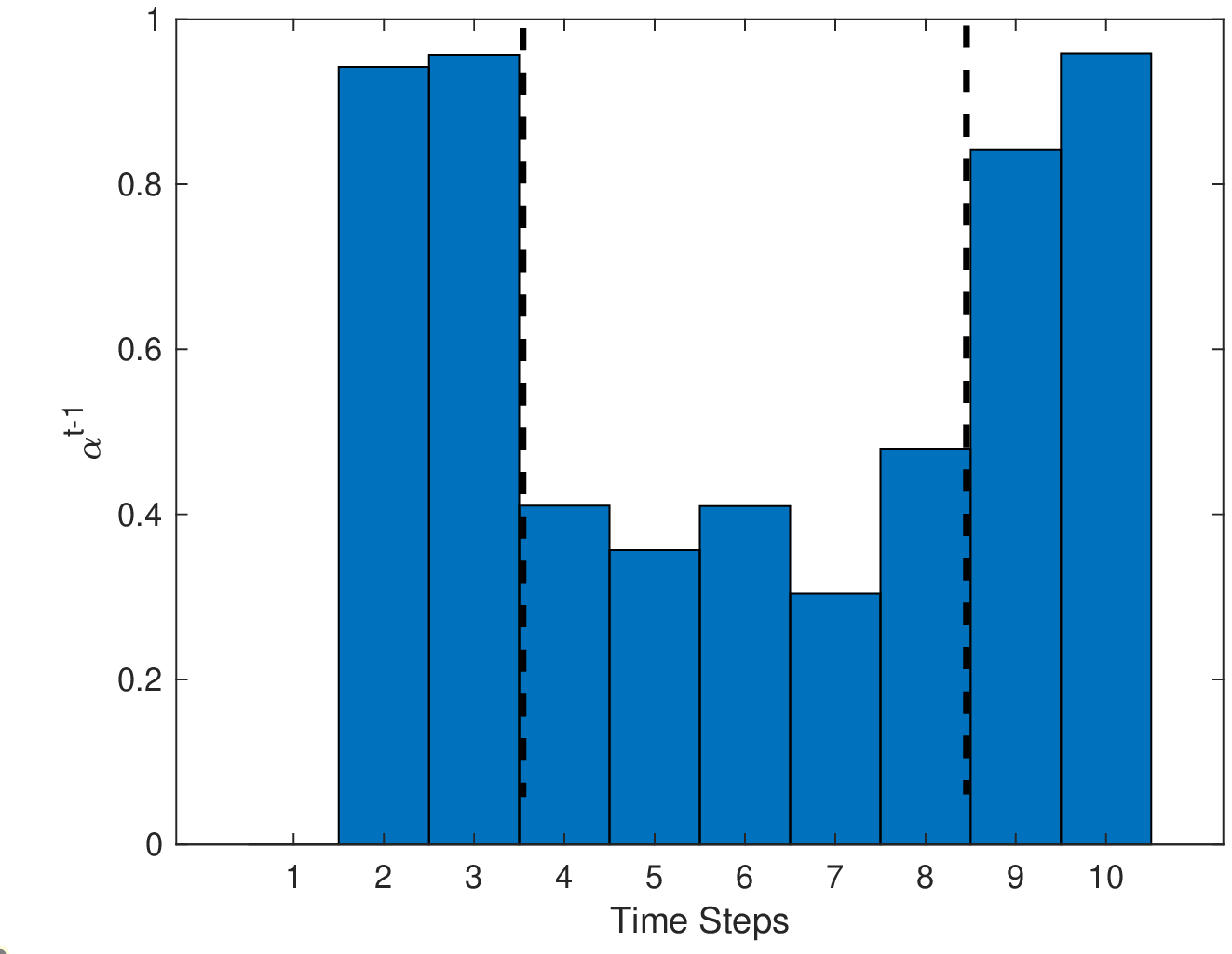}}
	\caption{Average weight of temporal smoothness among all runs of ERCOT at each time step of (a) Syn3 and (b) Syn7.}
\end{figure}

Syn3 is a challenging dataset with concept drift. We display the weight of temporal smoothness obtained by ERCOT at each time of this dataset in Figure \ref{fig_w_syn3}. The smoothness weights in Figure \ref{fig_w_syn3} correctly show the relevance between data at successive time steps: The weight goes lower from time 4 because the concept drift and cluster overlapping make the data prototypes at adjacent time steps be dissimilar. Neither concept drift nor cluster overlapping occurs since time 8; thus, the weight gets back to a higher value at time 9. The large weights in times 2, 3, 9, and 10 are reasonable: The data are normalized to zero mean before inferring the weight, therefore the previous data (in each cluster) can play an essential role in revealing the true latent probability density of the current data. Figure \ref{fig_w_syn7} exhibits the weight of temporal smoothness obtained by ERCOT at each time of Syn7. Similar to the performance in Syn3, the weights properly show the relevance between data at adjacent time periods even though the number of clusters in Syn7 varies over time. These observations demonstrate the feasibility and efficiency of the proposed smoothness weight auto-tuning method in ERCOT. 

\subsection{Time Efficiency of ERCOT}
We assess the running time of each algorithm in both synthetic and real data (detailed in Section \ref{sec_appliction}) on a PC with i7-6700 3.40 GHz CPU and 8G RAM. Results are reported in Table \ref{tabtime}.

\begin{table}[t]
	\caption{Average running time (seconds) among all runs of each algorithm at one time step of Syn1, bird\_flocks, and video\_seg\_MO.}
	\centering
	\label{tabtime}	
	\begin{tabular}{p{2cm}p{1.5cm}p{1.5cm}p{2cm}}
		\hline
		& Syn1   & bird\_flocks & video\_seg\_MO \\ \hline
		\emph{k}-meansCOT & 0.0069 & n/a    & n/a \\
		AFFECT            & 0.0603 & 0.0277 & n/a \\
		E$^2$SC           & 0.2296 & 0.0854 & 33.2716 \\
		penaltyEC         & 0.3525 & 0.2779 & 3.1248 \\
		staticEC          & 0.2653 & 0.2131 & 1.2170 \\
		ERCOT             & 0.2769 & 0.2748 & 1.9953 \\ \hline
	\end{tabular}
\end{table}

According to Table \ref{tabtime}, although ERCOT is slower than \emph{k}-meansCOT\footnote{We assume numbers of clusters of real datasets bird\_flocks and video\_seg\_MO are unknown a priori. \emph{k}-meansCOT requires predefining the number of clusters, thereby being unavailable for these datasets.} and AFFECT, the time it consumes is acceptable concerning to E$^2$SC, penaltyEC, and staticEC. Particularly, video\_seg\_MO is a large-scale dataset with 19200 samples. AFFECT fails to solve this dataset due to the unavailable memory demand of the affinity matrix. E$^2$SC is time-consuming in video\_seg\_MO because of the heavy computational load of solving the eigenvalue problem \cite{langone2016efficient}. By contrast, ERCOT requires much less running time, illustrating its time efficiency. 

Furthermore, staticEC is the static version of ERCOT without considering temporal smoothness. The running time of ERCOT is very close to those of staticEC, validating that the smoothness weight auto-tuning in ERCOT incurs insignificant computational burden to the algorithm. 

\begin{figure}[t] 
	\centering
	\includegraphics[width=0.9\linewidth]{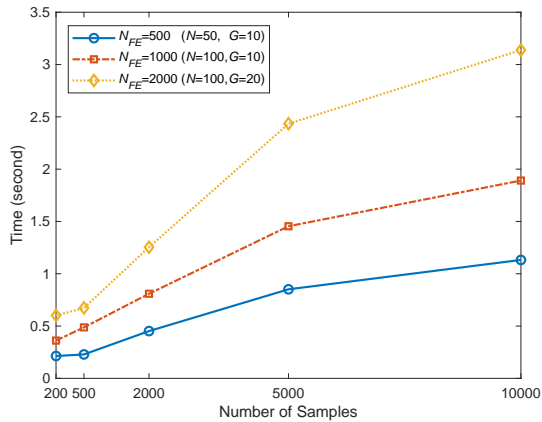}
	\caption{Running time (seconds) of ERCOT with different fitness evaluations versus the number of samples at one time step. $N$ is population size, $G$ is the number of generations, and $N_{EF}=N*G$ is the number of fitness evaluations.}
	\label{figTime}
\end{figure}

Additionally, we investigate the scalability of ERCOT in terms of time efficiency. The results in Figure \ref{figTime} show that the running time of ERCOT increases roughly linearly with the increase in both fitness evaluations and the number of samples. The above results demonstrate the time efficiency of ERCOT.  

\subsection{Applications}\label{sec_appliction}
\subsubsection{Bird Flocks Tracking}
\begin{figure}[t] 
	\centering
	\includegraphics[width=0.9\linewidth]{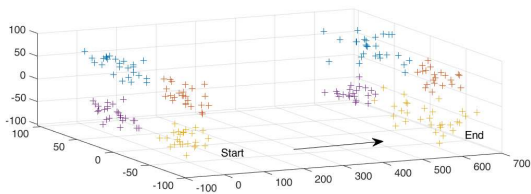}
	\caption{Illustration of bird\_flocks dataset. Four flocks fly along parallel paths. At each time step, concept drift happens when stray bird joins one of the other flocks.}
	\label{figboid}
\end{figure}

The bird flocks tracking simulates the natural movement and flocking behavior of birds. According to the model in \cite{reynolds1987flocks}, birds' behavior obeys three rules: 1) flying towards the center of the flock it belongs to; 2) keeping a small distance away from other birds; and 3) trying to match velocity with near birds. We produce the bird\_flocks dataset (Figure \ref{figboid}) similar to that in \cite{xu2014adaptive}\cite{parkerboids2007}. Initially, four flocks, each with 25 uniformly distributed birds, are drawn. At each time step, birds move following the above three rules; and a stray bird of each flock will join one of the other flocks. Totally 30 time steps are considered. The actual number of clusters is set to be unknown a priori in the experiment on this dataset.
  
\begin{figure}[t] 
	\centering
	\includegraphics[width=0.9\linewidth]{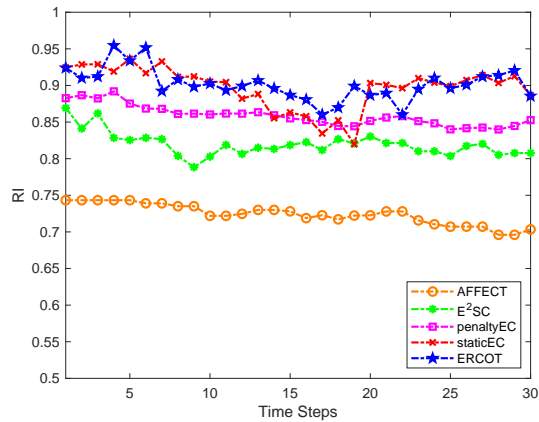}
	\caption{Average RI value among all runs of each algorithm at each time step of bird\_flocks.}
	\label{figRIboid}
\end{figure}

The RI results of algorithms versus time steps are plotted in Figure \ref{figRIboid}. The mRI among all time steps is given in Table \ref{tabRIboid}. \emph{k}-meansCOT is unavailable in these datasets because it requires predefining the number of clusters. AFFECT, E$^2$SC, and penaltyEC are inferior to the static approach staticEC, possibly because they fail to determine a correct number of clusters and struggle to predefine a reasonable preference to the temporal smoothness. Compared with staticEC in Figure \ref{figRIboid}, ERCOT performs more stable and continually achieves promising results over time, demonstrating its validity.    

\begin{table}[t]
	\caption{Mean and standard deviation of mRI results of bird flocks tracking. Best values are bold.}
	\centering
	\label{tabRIboid}
	\begin{threeparttable}	
		\begin{tabular}{p{3cm}p{5cm}}
			\hline
			Algorithm         & mRI                    \\ \hline
			\emph{k}-meansCOT & n/a                    \\
			AFFECT            & 0.7099$\pm$4.01E-02$>$ \\
			E$^2$SC           & 0.8144$\pm$0.00E+00$>$ \\
			penaltyEC         & 0.8501$\pm$1.20E-03$>$ \\
			staticEC          & 0.8930$\pm$1.18E-02$\approx$ \\
			ERCOT             & \textbf{0.8951}$\pm$5.50E-03 \\ \hline
		\end{tabular}
		\begin{tablenotes}
		\scriptsize
		\item $>$: The performance of ERCOT is significantly better than the outcome according to the Wilcoxon Sign Test at the 0.05 significance level.
		\item $\approx$: The performance of ERCOT is comparable with the outcome according to the Wilcoxon Sign Test at the 0.05 significance level.
	    \end{tablenotes}
    \end{threeparttable}
\end{table}

\subsubsection{Video Background Segmentation}
\begin{figure}[t]
	\centering
	\subfigure[]{\includegraphics[width=1.05in,height=0.85in]{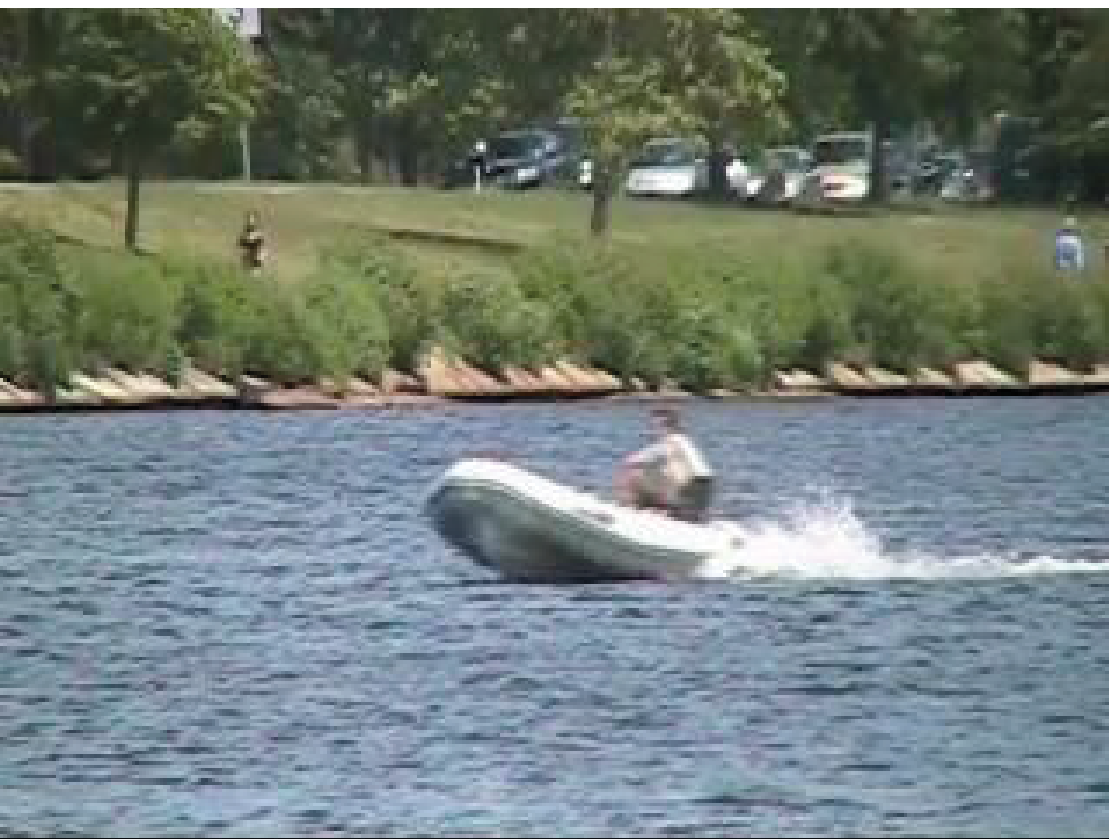}}
	\subfigure[]{\includegraphics[width=1.05in,height=0.85in]{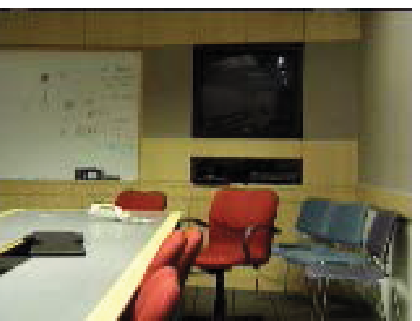}}
	\subfigure[]{\includegraphics[width=1.05in,height=0.85in]{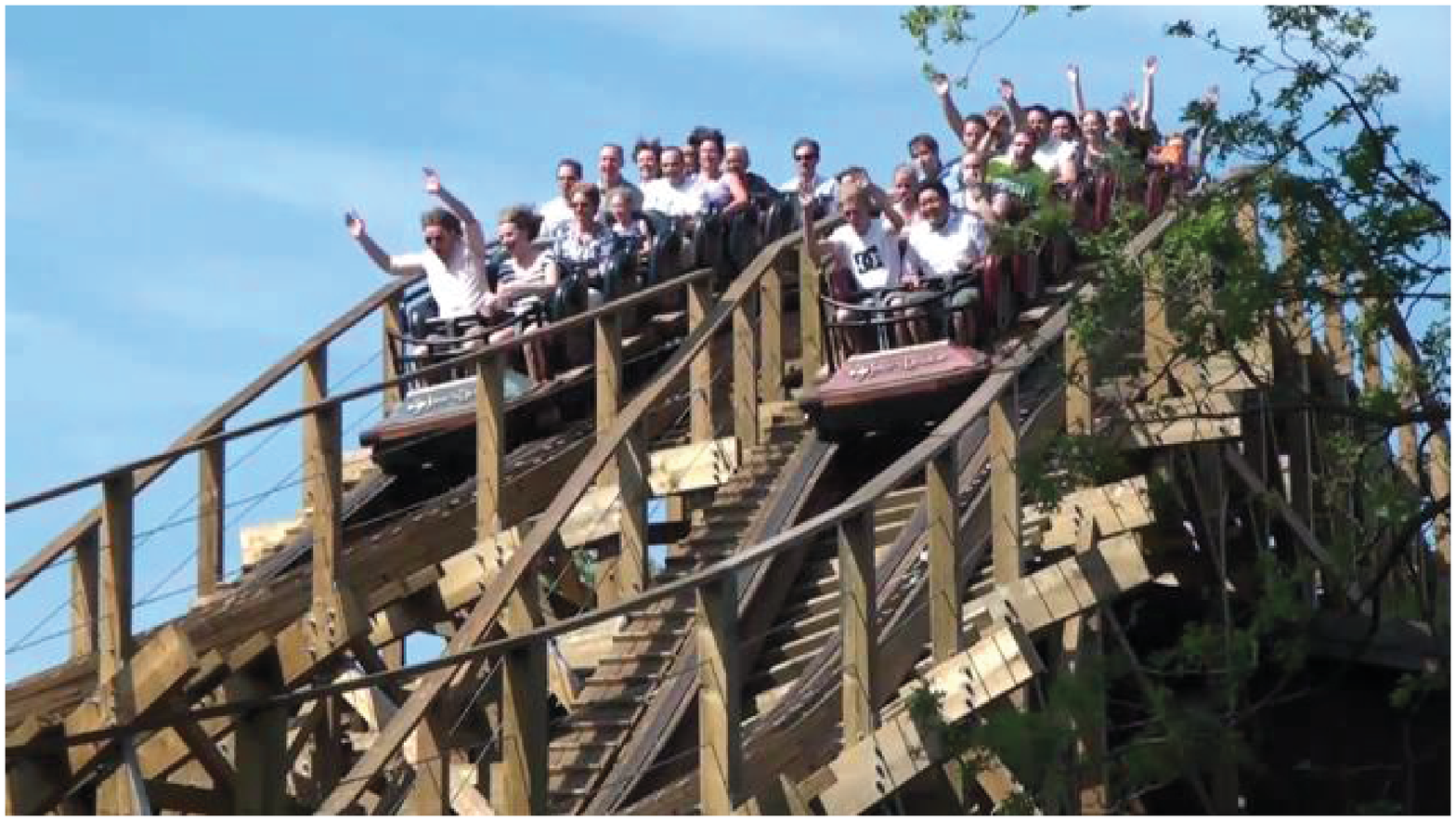}}
	\caption{Scenarios of video sequences (a) video\_seg\_BT, (b) video\_seg\_MO, and (c) video\_seg\_RC.}
	\label{fig_video_org}
\end{figure}

Video background segmentation refers to grouping perceptually similar pixels into regions at each frame of a video. The region segmentations should keep coherence in the time-variant scenarios due to the temporal smooth nature of scenes \cite{chacon2018bio}. 
We experiment with three video segmentation tasks with different characteristics, as shown in Figure \ref{fig_video_org}. Video\_seg\_BT \cite{BT} has dynamic backgrounds with static objects. It consists of 12 frames, each with $320\times240$ pixels, and the number of segmentations (clusters) varies between 4 and 5. Video\_seg\_MO \cite{MO} has a static object that changes its position. It involves 9 frames, each with $160\times120$ pixels and 7 segmentations. Finally, video\_seg\_RC \cite{galasso2013unified} has a snailing background with moving objects. It contains 8 frames, each with $640\times360$ pixels, and the number of segmentations varies between 3 and 4. The three datasets are collected in the RGB color space. The actual number of clusters is assumed to be unknown a priori.

The mRI performance of algorithms is given in Table \ref{tabRIimg}. \emph{k}-meansCOT and AFFECT cannot handle these datasets. The former requires predefining the number of clusters, while the latter's heavy memory demand of the affinity matrix is unavailable in these large-scale datasets (implemented by the AFFECT Matlab toolbox \cite{xu2014adaptive}). The proposed ERCOT obtains the best results in two datasets, and the value it reached in the rest dataset is competitive concerning other algorithms. E$^2$SC performs well in Video\_seg\_BT and Video\_seg\_RC, showing its efficiency. 

\begin{figure}[t] 
	\centering
    \includegraphics[width=1\linewidth]{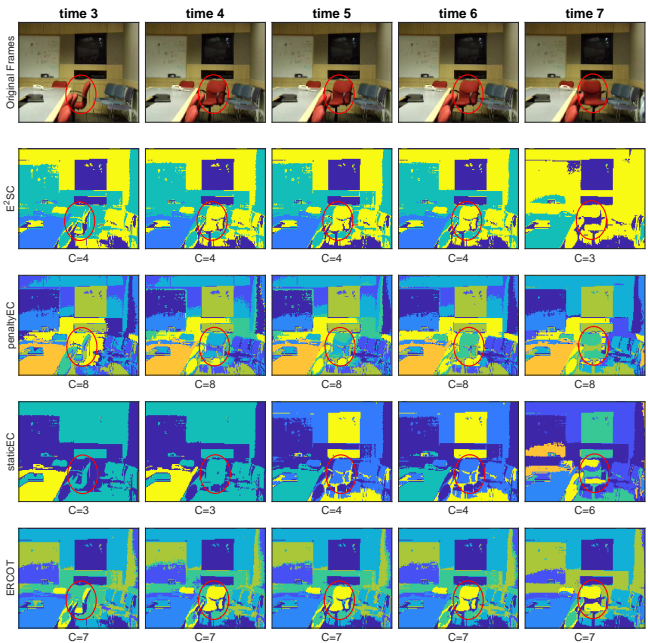}
	\caption{Original frames and algorithms' segmentations of time 3-7 of video\_seg\_MO. C is the number of clusters. Red ellipses mark the object whose position changes over time.}
	\label{figsegmo}
\end{figure}

The segmentation results of video\_seg\_MO are exhibited in Figure \ref{figsegmo}. These results visualize algorithms' ability to temporal smoothness maintenance. According to Figure \ref{figsegmo}, penaltyEC always identifies larger numbers of clusters than the true numbers because its clustering validity index $J_{cp}$ favors partitions with a huge number of clusters. StaticEC clusters each frame independently, hence cannot keep segmentations smooth in successive frames. Furthermore, its clustering quality in each frame is inferior to other algorithms, and it fails to determine a proper number of clusters. The reason is that for these large-scale datasets, it cannot converge well within the limited fitness evaluations in the absence of historical information's assistance. In comparison, the proposed ERCOT tracks the rotating chair (marked within red ellipses in Figure \ref{figsegmo}) over time. This promising performance benefits from 1) promoting convergence by utilizing historical information in the reinitialization step (lines 4-7 of Algorithm \ref{algWorkflow}) and 2) maintaining temporal smoothness by fitness accumulation (lines 9-12 of Algorithm \ref{algWorkflow}). Additionally, ERCOT identifies the actual number of segmentations in each frame. The above results demonstrate ERCOT's practicability. 

\begin{table*}[t]
	\caption{Mean and standard deviation of mRI results of video background segmentation. Best values are bold.}
	\centering
	\label{tabRIimg}
	\begin{threeparttable}
		\begin{tabular}{lllllll}
			\hline
			& \multicolumn{1}{c}{\emph{k}-meansCOT} & \multicolumn{1}{c}{AFFECT} & \multicolumn{1}{c}{E$^2$SC} & \multicolumn{1}{c}{penaltyEC} & \multicolumn{1}{c}{staticEC} & \multicolumn{1}{c}{ERCOT}  \\ \hline
			video\_seg\_BT & n/a & n/a & \textbf{0.7514}$\pm$0.00E+00 & 0.7042$\pm$6.80E-03$>$ & 0.7157$\pm$8.50E-03$>$ & 0.7335$\pm$4.60E-03 \\
			video\_seg\_MO & n/a & n/a & 0.6706$\pm$0.00E+00$>$ & 0.6952$\pm$1.30E-03$>$ & 0.6399$\pm$4.70E-03$>$ & \textbf{0.7163}$\pm$6.00E-04 \\
			video\_seg\_RC & n/a & n/a & 0.7499$\pm$0.00E+00$\approx$ & 0.7366$\pm$6.20E-03$>$ & 0.7297$\pm$2.35E-02$>$ & \textbf{0.7526}$\pm$1.55E-02 \\
			\hline
		\end{tabular}
		\begin{tablenotes}
			\scriptsize
			\item $>$: The performance of ERCOT is significantly better than the outcome according to the Wilcoxon Sign Test at the 0.05 significance level.
			\item $\approx$: The performance of ERCOT is comparable with the outcome according to the Wilcoxon Sign Test at the 0.05 significance level.
		\end{tablenotes} 
	\end{threeparttable}
\end{table*}

\section{Conclusions}\label{secConclusion}
This paper proposed a novel partitional clustering framework ERCOT for temporal data. Unlike existing algorithms, ERCOT processes the temporal smoothness in an \emph{a posterior} way, avoiding the potential unexpected convergence in current methods. Furthermore, it automatically tunes the weight of smoothness by quantifying the significance of previous data on the current clustering; and it automatically determines the number of clusters via optimizing multiple partitions at different hierarchical levels in a single run. These features enable ERCOT to hold better scalability and practicality than that of existing algorithms. The comprehensive and extensive experimental analysis demonstrated the superiority of ERCOT over existing algorithms. 

The proposed ERCOT involves cluster prototype-based objective function and centroid-based encoding, which is fitter for sphere clusters. In the future, we plan to incorporate cluster label-based objective functions and design advanced encoding schemes for distinctively shaped datasets. Furthermore, it is desirable to equip ERCOT with problem-specific (e.g., dynamic community detection, video segmentation, etc.) techniques for better practical problem-solving, which are also our future research topics.

\bibliographystyle{IEEEtran}
\bibliography{References}

\begin{thebibliography}{10}
\providecommand{\url}[1]{#1}
\csname url@samestyle\endcsname
\providecommand{\newblock}{\relax}
\providecommand{\bibinfo}[2]{#2}
\providecommand{\BIBentrySTDinterwordspacing}{\spaceskip=0pt\relax}
\providecommand{\BIBentryALTinterwordstretchfactor}{4}
\providecommand{\BIBentryALTinterwordspacing}{\spaceskip=\fontdimen2\font plus
\BIBentryALTinterwordstretchfactor\fontdimen3\font minus
  \fontdimen4\font\relax}
\providecommand{\BIBforeignlanguage}[2]{{%
\expandafter\ifx\csname l@#1\endcsname\relax
\typeout{** WARNING: IEEEtran.bst: No hyphenation pattern has been}%
\typeout{** loaded for the language `#1'. Using the pattern for}%
\typeout{** the default language instead.}%
\else
\language=\csname l@#1\endcsname
\fi
#2}}
\providecommand{\BIBdecl}{\relax}
\BIBdecl

\bibitem{xu2014adaptive}
K.~S. Xu, M.~Kliger, and A.~O. Hero~Iii, ``Adaptive evolutionary clustering,''
  \emph{Data Mining Knowl. Discovery}, vol.~28, no.~2, pp. 304--336, Jan. 2014.

\bibitem{chakrabarti2006evolutionary}
D.~Chakrabarti, R.~Kumar, and A.~Tomkins, ``Evolutionary clustering,'' in
  \emph{proc. Int. Conf. Knowl. Discovery Data Mining}, Philadelphia, PA, USA,
  2006, pp. 554--560.

\bibitem{chacon2018bio}
M.~I. Chacon-Murguia and J.~A. Ramirez-Quintana, ``Bio-inspired architecture
  for static object segmentation in time varying background models from video
  sequences,'' \emph{Neurocomputing}, vol. 275, pp. 1846--1860, Jan. 2018.

\bibitem{xia2018human}
G.~Xia, H.~Sun, L.~Feng, G.~Zhang, and Y.~Liu, ``Human motion segmentation via
  robust kernel sparse subspace clustering,'' \emph{IEEE Trans. Image Proc.},
  vol.~27, no.~1, pp. 135--150, Jan. 2018.

\bibitem{folino2014evolutionary}
F.~Folino and C.~Pizzuti, ``An evolutionary multiobjective approach for
  community discovery in dynamic networks,'' \emph{IEEE Trans. Knowl. Data
  Eng.}, vol.~26, no.~8, pp. 1838--1852, Aug. 2014.

\bibitem{bara2016new}
A.~A. Bara'a and H.~S. Khoder, ``A new multi-objective evolutionary framework
  for community mining in dynamic social networks,'' \emph{Swarm Evol.
  Comput.}, vol.~31, pp. 90--109, Dec. 2016.

\bibitem{zeng2019consensus}
X.~Zeng, W.~Wang, C.~Chen, and G.~G. Yen, ``A consensus community-based
  particle swarm optimization for dynamic community detection,'' \emph{IEEE
  Trans. Cyber.}, vol.~50, no.~6, pp. 2502--2513, Jun. 2020.

\bibitem{rana2014evolutionary}
C.~Rana and S.~K. Jain, ``An evolutionary clustering algorithm based on
  temporal features for dynamic recommender systems,'' \emph{Swarm Evol.
  Comput.}, vol.~14, pp. 21--30, Feb. 2014.

\bibitem{lin2019opec}
R.~Lin, Z.~Ye, and Y.~Zhao, ``Opec: Daily load data analysis based on optimized
  evolutionary clustering,'' \emph{Energies}, vol.~12, no.~14, p. 2668, Jul.
  2019.

\bibitem{chen2014evolutionary}
G.~Chen, W.~Luo, and T.~Zhu, ``Evolutionary clustering with differential
  evolution,'' in \emph{proc. IEEE Congr. Evol. Comput.}\hskip 1em plus 0.5em
  minus 0.4em\relax Beijing, China: IEEE, 2014, pp. 1382--1389.

\bibitem{li2015sparsity}
R.~Li, W.~Zhang, Y.~Zhao, Z.~Zhu, and S.~Ji, ``Sparsity learning formulations
  for mining time-varying data,'' \emph{IEEE Trans. Knowl. Data Eng.}, vol.~27,
  no.~5, pp. 1411--1423, May 2015.

\bibitem{langone2016efficient}
R.~Langone, M.~Van~Barel, and J.~A. Suykens, ``Efficient evolutionary spectral
  clustering,'' \emph{Pattern Recognit. Lett.}, vol.~84, pp. 78--84, Dec. 2016.

\bibitem{wang2017hierarchical}
P.~Wang, P.~Zhang, C.~Zhou, Z.~Li, and H.~Yang, ``Hierarchical evolving
  dirichlet processes for modeling nonlinear evolutionary traces in temporal
  data,'' \emph{Data Mining Knowl. Discovery}, vol.~31, no.~1, pp. 32--64, Feb.
  2017.

\bibitem{chi2009evolutionary}
Y.~Chi, X.~Song, D.~Zhou, K.~Hino, and B.~L. Tseng, ``On evolutionary spectral
  clustering,'' \emph{ACM Trans. Knowl. Discovery Data}, vol.~3, no.~4, pp.
  1--30, Nov. 2009.

\bibitem{hashemi2018evolutionary}
A.~Hashemi and H.~Vikalo, ``Evolutionary self-expressive models for subspace
  clustering,'' \emph{IEEE J. Sel. Top. Sign. Proc.}, vol.~12, no.~6, pp.
  1534--1546, Dec. 2018.

\bibitem{arzeno2017evolutionary}
N.~M. Arzeno and H.~Vikalo, ``Evolutionary affinity propagation,'' in
  \emph{proc. IEEE Int. Conf. Acoust. Speech Sign. Proc.}\hskip 1em plus 0.5em
  minus 0.4em\relax New Orleans, LA, USA: IEEE, 2017, pp. 2681--2685.

\bibitem{arzeno2019evolutionary}
------, ``Evolutionary clustering via message passing,'' \emph{IEEE Trans.
  Knowl. Data Eng.}, to be published, DOI: 10.1109/TKDE.2019.2954869.

\bibitem{strehl2002cluster}
A.~Strehl and J.~Ghosh, ``Cluster ensembles---a knowledge reuse framework for
  combining multiple partitions,'' \emph{J. Mach. Learn. Res.}, vol.~3, pp.
  583--617, Dec. 2002.

\bibitem{li2020does}
K.~Li, M.~Liao, K.~Deb, G.~Min, and X.~Yao, ``Does preference always help? a
  holistic study on preference-based evolutionary multi-objective optimisation
  using reference points,'' \emph{IEEE Trans. Evol. Comput.}, vol.~24, no.~6,
  pp. 1078--1096, Dec. 2020.

\bibitem{fu2015robust}
H.~Fu, B.~Sendhoff, K.~Tang, and X.~Yao, ``Robust optimization over time:
  Problem difficulties and benchmark problems,'' \emph{IEEE Trans. Evol.
  Comput.}, vol.~19, no.~5, pp. 731--745, Oct. 2015.

\bibitem{hartigan1979algorithm}
J.~A. Hartigan and M.~A. Wong, ``Algorithm as 136: A k-means clustering
  algorithm,'' \emph{J. Royal Stat. Society (Applied Stat.)}, vol.~28, no.~1,
  pp. 100--108, 1979.

\bibitem{ma2011spatio}
J.~Ma, Y.~Wang, M.~Gong, L.~Jiao, and Q.~Zhang, ``Spatio-temporal data
  evolutionary clustering based on moea/d,'' in \emph{proc. Int. Conf. Gene.
  Evol. Comput.}, Dublin, Ireland, 2011, pp. 85--86.

\bibitem{chen2015clustering}
G.~Chen and W.~Luo, ``Clustering time-evolving data using an efficient
  differential evolution,'' in \emph{proc. Int. Conf. Swarm Intel.}\hskip 1em
  plus 0.5em minus 0.4em\relax Beijing, China: Springer, 2015, pp. 326--338.

\bibitem{gao2016clustering}
W.~Gao, W.~Luo, C.~Bu, L.~Ni, and D.~Zhang, ``Clustering evolutionary data with
  an r-dominance based multi-objective evolutionary algorithm,'' in \emph{proc.
  Int. Conf. Intel. Data Eng. Auto. Learn.}\hskip 1em plus 0.5em minus
  0.4em\relax Springer, 2016, pp. 342--352.

\bibitem{liu2019adaptive}
C.~Liu, Q.~Zhao, B.~Yan, S.~Elsayed, T.~Ray, and R.~Sarker, ``Adaptive
  sorting-based evolutionary algorithm for many-objective optimization,''
  \emph{IEEE Trans. Evol. Comput.}, vol.~23, no.~2, pp. 247--257, Apr. 2019.

\bibitem{zhao2021evolutionary}
Q.~Zhao, B.~Yan, Y.~Shi, and M.~Middendorf, ``Evolutionary dynamic
  multiobjective optimization via learning from historical search process,''
  \emph{IEEE Trans. Cyber.}, to be published, DOI: 10.1109/TCYB.2021.3059252.

\bibitem{mukhopadhyay2015survey}
A.~Mukhopadhyay, U.~Maulik, and S.~Bandyopadhyay, ``A survey of multiobjective
  evolutionary clustering,'' \emph{ACM Comput. Surveys}, vol.~47, no.~4, pp.
  1--46, May 2015.

\bibitem{liu2019transfer}
C.~Liu, Q.~Zhao, B.~Yan, S.~Elsayed, and R.~Sarker, ``Transfer
  learning-assisted multi-objective evolutionary clustering framework with
  decomposition for high-dimensional data,'' \emph{Informat. Sci.}, vol. 505,
  pp. 440--456, Dec. 2019.

\bibitem{deb2007dynamic}
K.~Deb, S.~Karthik \emph{et~al.}, ``Dynamic multi-objective optimization and
  decision-making using modified nsga-ii: a case study on hydro-thermal power
  scheduling,'' in \emph{proc. Int. Conf. Evol. Multi-criterion Opti.},
  Matsushima, Japan, 2007, pp. 803--817.

\bibitem{deb1995simulated}
K.~Deb, R.~B. Agrawal \emph{et~al.}, ``Simulated binary crossover for
  continuous search space,'' \emph{Complex Syst.}, vol.~9, no.~2, pp. 115--148,
  Nov. 1995.

\bibitem{deb1996combined}
K.~Deb and M.~Goyal, ``A combined genetic adaptive search (geneas) for
  engineering design,'' \emph{Computer Sci. Informatics}, vol.~26, pp. 30--45,
  1996.

\bibitem{deb2002fast}
K.~Deb, A.~Pratap, S.~Agarwal, and T.~Meyarivan, ``A fast and elitist
  multiobjective genetic algorithm: Nsga-ii,'' \emph{IEEE Trans. Evol.
  Comput.}, vol.~6, no.~2, pp. 182--197, Apr. 2002.

\bibitem{handl2007evolutionary}
J.~Handl and J.~Knowles, ``An evolutionary approach to multiobjective
  clustering,'' \emph{IEEE Trans. Evol. Comput.}, vol.~11, no.~1, pp. 56--76,
  Feb. 2007.

\bibitem{garza2018improved}
M.~Garza-Fabre, J.~Handl, and J.~Knowles, ``An improved and more scalable
  evolutionary approach to multiobjective clustering,'' \emph{IEEE Trans. Evol.
  Comput.}, vol.~22, no.~4, pp. 515--535, Aug. 2018.

\bibitem{bezdek1981pattern}
J.~C. Bezdek, \emph{Pattern Recognition with Fuzzy Objective Function
  Algorithms}.\hskip 1em plus 0.5em minus 0.4em\relax New York, USA: Plenum,
  1981.

\bibitem{hancer2017comprehensive}
E.~Hancer and D.~Karaboga, ``A comprehensive survey of traditional, merge-split
  and evolutionary approaches proposed for determination of cluster number,''
  \emph{Swarm Evol. Comput.}, vol.~32, pp. 49--67, Feb. 2017.

\bibitem{smyth1997stacked}
P.~Smyth and D.~Wolpert, ``Stacked density estimation,'' in \emph{proc. Neural
  Informat. Proc. Syst.}, Denver, CO, USA, 1997, pp. 668--674.

\bibitem{da2019curbing}
B.~Da, A.~Gupta, and Y.-S. Ong, ``Curbing negative influences online for
  seamless transfer evolutionary optimization,'' \emph{IEEE Trans. Cyber.},
  vol.~49, no.~12, pp. 4365--4378, Dec. 2019.

\bibitem{kuhn1955hungarian}
H.~W. Kuhn, ``The hungarian method for the assignment problem,'' \emph{Naval
  Res. Logist. Q.}, vol.~2, no. 1-2, pp. 83--97, Mar. 1955.

\bibitem{moon1996expectation}
T.~K. Moon, ``The expectation-maximization algorithm,'' \emph{IEEE Sign. Proc.
  Mag.}, vol.~13, no.~6, pp. 47--60, Nov. 1996.

\bibitem{reynolds1987flocks}
C.~W. Reynolds, ``Flocks, herds and schools: A distributed behavioral model,''
  in \emph{proc. Conf. Comput. Graph. Interact. Tech.}, 1987, pp. 25--34.

\bibitem{rand1971objective}
W.~M. Rand, ``Objective criteria for the evaluation of clustering methods,''
  \emph{J. Am. Stat. Assoc.}, vol.~66, no. 336, pp. 846--850, 1971.

\bibitem{d1986principles}
R.~G. d~Steel and J.~H. Torrie, \emph{Principles and procedures of statistics:
  A biometrical approach}.\hskip 1em plus 0.5em minus 0.4em\relax Auckland, New
  Zealand: McGraw-Hill, 1986.

\bibitem{rousseeuw1987silhouettes}
P.~J. Rousseeuw, ``Silhouettes: a graphical aid to the interpretation and
  validation of cluster analysis,'' \emph{J. Comput. Applied Math.}, vol.~20,
  pp. 53--65, 1987.

\bibitem{parkerboids2007}
C.~Parker, ``Boids pseudocode,'' \emph{URL
  http://www.kfish.org/boids/pseudocode.html}.

\bibitem{BT}
``Changedetection.net(cdnet): A video database for testing change detection
  algorithms,'' \emph{URL http://www.changedetection.net/}.

\bibitem{MO}
``Test images for wallflower paper,'' \emph{URL
  https://www.microsoft.com/en-us/download/details.aspx?id=54651}.

\bibitem{galasso2013unified}
F.~Galasso, N.~Shankar~Nagaraja, T.~Jimenez~Cardenas, T.~Brox, and B.~Schiele,
  ``A unified video segmentation benchmark: Annotation, metrics and analysis,''
  in \emph{proc. IEEE Int. Conf. Comput. Vision}, Sydney, Australia, 2013, pp.
  3527--3534.

\end{thebibliography}
\end{document}